\documentclass{ieeeaccess}

\usepackage{cite}
\usepackage{amsmath,amssymb,amsfonts}
\usepackage{algorithmic}
\usepackage{graphicx}
\usepackage{stfloats}
\usepackage{placeins}
\usepackage{textcomp}
\usepackage{bm}
\usepackage{array}
\usepackage[table]{xcolor}
\usepackage{soul}
\usepackage{url}
\usepackage{xurl}
\usepackage{amsmath}
\usepackage{xcolor}

\usepackage[caption=false,font=footnotesize]{subfig}

\soulregister\ref7
\soulregister\pageref7
\soulregister\cite7

\newcommand{\hleq}[1]{%
    \text{\colorbox{yellow}{$\displaystyle #1$}}%
}

\newif\ifshowhighlights

\showhighlightsfalse

\ifshowhighlights
\else
    \renewcommand{\hl}[1]{#1}
    
    \renewcommand{\hleq}[1]{#1}

    \renewcommand{\rowcolors}[3]{}
    \renewcommand{\rowcolor}[1]{}

\fi

\makeatletter
\AtBeginDocument{%
    \DeclareMathVersion{bold}
    \SetSymbolFont{operators}{bold}{T1}{times}{b}{n}
    \SetSymbolFont{NewLetters}{bold}{T1}{times}{b}{it}
    \SetMathAlphabet{\mathrm}{bold}{T1}{times}{b}{n}
    \SetMathAlphabet{\mathit}{bold}{T1}{times}{b}{it}
    \SetMathAlphabet{\mathbf}{bold}{T1}{times}{b}{n}
    \SetMathAlphabet{\mathtt}{bold}{OT1}{pcr}{b}{n}
    \SetSymbolFont{symbols}{bold}{OMS}{cmsy}{b}{n}
    \renewcommand\boldmath{\@nomath\boldmath\mathversion{bold}}%
}
\makeatother

\def\BibTeX{%
    {\rm B\kern-.05em{\sc i\kern-.025em b}\kern-.08em
    T\kern-.1667em\lower.7ex\hbox{E}\kern-.125emX}%
}
\begin{document}
\history{Date of publication xxxx 00, 0000, date of current version xxxx 00, 0000.}
\doi{10.1109/ACCESS.2026.0000000}

\title{A Novel Lightweight Multi-Scale Branching Encoder for Few-Shot Image Classification}

\author{\uppercase{Oriod Malo}\authorrefmark{1},
\uppercase{Alexandre Xavier Falc{\~a}o}\authorrefmark{2}, and
\uppercase{Laurent Najman}\authorrefmark{3,4}}

\address[1]{AI Laboratory, Faculty of Computer Science and Information Technology, University Metropolitan Tirana, Tirana, Albania (e-mail: omalo@umt.edu.al)}
\address[2]{Institute of Computing (IC), University of Campinas (UNICAMP), Campinas, Brazil (e-mail: afalcao@ic.unicamp.br)}
\address[3]{Math. Department, Khalifa University, Abu Dhabi, UAE}
\address[4]{LIGM, Universit{\'e} Gustave Eiffel, CNRS, Marne-la-Vall{\'e}e, France (e-mail: laurent.najman@esiee.fr)}

\markboth
{Malo \headeretal: A Novel Lightweight Multi-Scale Branching Encoder for Few-Shot Image Classification}
{Malo \headeretal: A Novel Lightweight Multi-Scale Branching Encoder for Few-Shot Image Classification}

\corresp{Corresponding author: Laurent Najman (e-mail: laurent.najman@esiee.fr).}

\begin{abstract}
Few-Shot Learning (FSL) addresses classification scenarios in which only a small number of labeled examples are available for each class. This paper presents the Sum12 Multi-Scale branching encoder, a lightweight convolutional architecture designed to improve the embedding stage of Prototypical Networks without substantially increasing computational cost. The encoder uses seven parallel convolutional branches whose kernel-size combinations sum to 12, allowing local and broader spatial structures to be represented within a unified Multi-Scale feature space. The proposed architecture was evaluated on Omniglot, CIFAR-FS, FC100, and miniImageNet under controlled few-shot learning protocols and compared with the original Prototypical Network encoder and Inception-inspired branch variants. Additional ablation studies have investigated fusion strategies, dilated convolutions, branch compositions, and receptive field budget. Beyond few-shot learning, the encoder was also tested on the MedMNIST 2D benchmark to assess its behavior in conventional supervised medical image classification. \hl{Across these experiments, Sum12 generally improved the low-shot classification performance with respect to the basic ProtoNet encoder while preserving a compact model size.} \hl{These properties suggest that the proposed encoder may be suitable for low-data visual recognition and resource-constrained intelligent imaging applications. However, complete end-to-end validation on the target embedded platform, including resolution of the observed implementation discrepancies, remains future work.}
\end{abstract}


\titlepgskip=-21pt
\maketitle
\section{Introduction}

\PARstart{T}{extbook} Machine Learning and Deep Learning solutions are typically developed under the assumption that sufficient data is available for the network to effectively learn the target concept. However, many real-life scenarios violate this assumption, particularly in medical imaging, where data heterogeneity, annotation costs, and variability across acquisition settings limit the availability of labeled samples. The branch of Machine Learning that specializes in training models with few data points is known as Few-Shot Learning (FSL), which provides a natural framework for studying learning under severe data constraints.

Despite extensive progress in Few-Shot Learning, most studies have focused on meta-learning strategies, metric learning formulations, and training paradigms. Simultaneously, the design of the encoder, which is the fundamental feature extractor that forms the base of these systems, has remained comparatively simple. In particular, Prototypical Networks, one of the most influential and broadly adopted FSL methods, rely on a basic convolutional encoder, which limits their ability to extract diverse, scale-invariant representations. This limitation is particularly evident in image analysis problems in which relevant structures occur at multiple spatial scales, which is a common characteristic of medical images. This design limitation motivates the search for a more expressive encoder that enhances the feature diversity without impairing the computational efficiency.

To address this limitation, the Sum12 Multi-Scale encoder is introduced as a novel, lightweight convolutional architecture designed to enhance representation learning in data-limited and scale-heterogeneous environments. This encoder was evaluated in the Few-Shot Learning framework using a controlled and established testing protocol. The architecture employs a branching multi-scale configuration that combines convolutional paths with varying kernel-size sequences, each constrained to a cumulative kernel-size sum of 12. This approach enables the network to capture both local and global patterns within a single feature-extraction stage, resulting in a more robust and semantically rich embedding space.

We extensively evaluated the proposed encoder on multiple standard Few-Shot benchmarks such as Omniglot, CIFAR-FS, FC100, and miniImageNet, and compared it with the baseline Prototypical Networks. Furthermore, we benchmarked its behavior against the Inception module \cite{szegedy2015googlenet}, which similarly employs branching convolutions with varying kernel sizes, to better contextualize the proposed architecture within existing Multi-Scale designs.

Following the preliminary evaluation of the encoder, a series of ablation studies was conducted to investigate the effects of Multi-Scale branching, fusion strategies, kernel dilation, and branch composition under various training and evaluation conditions. These studies aimed to provide a deeper understanding of the behavior of the architecture in different contexts.

To assess generalization beyond Few-Shot Learning and evaluate the relevance of the proposed encoder for medical image analysis, the Sum12 Multi-Scale encoder was also tested on the MedMNIST 2D benchmark. This benchmark comprises 12 heterogeneous medical imaging datasets spanning multiple modalities and anatomical regions. 

\hl{The results in MedMNIST 2D demonstrate the eligibility of the encoder as a possible lightweight and high-performance model for conventional image classification tasks. In addition, the trained encoders remained relatively lightweight in size and relied exclusively on convolutional building blocks, making the architecture a suitable candidate for deployment studies in resource-constrained environments. Because this size exceeded the internal memory available on the microcontroller boards at our disposal, we deployed a quantized variant with dilated convolutions, which fit within the memory budget of the target board. However, the validation of the generated on-target implementation showed low agreement with the reference Python implementation, with approximately 35.2\% cosine similarity and 28.2\% mean absolute error. Therefore, the on-target implementation has not yet been functionally validated for few-shot classification.}

The remainder of this paper is organized as follows. Section~\ref{sec:theoretical_background} reviews the related work and outlines the theoretical background that supports this proposal. Section~\ref{sec:architecture} details the architectural design of the encoder, and Section~\ref{sec:experimental_setup} presents the experimental setup used to obtain the results reported in Section~\ref{sec:results_discussion}. Section~\ref{sec:conclusion} concludes the paper and discusses potential directions to further improve the Sum12 Multi-Scale branching encoder.

The first known FSL methodology used two identical convolutional neural networks (CNNs) with shared weights, trained in pairs labeled "same" or "different", thereby relying on the similarity between the input and the query \cite{koch2015}. Immediately thereafter, \cite{vinyals2016} incorporated attention mechanisms, an episodic training paradigm, and full-context embeddings into the Few-Shot Learning framework. The episodic training procedure enabled the model to train under conditions identical to those at test time, thereby supporting the classification with a dynamically varying support set. The full-context embedding is formulated such that a function conditioned on the entire set, excluding the sample itself, produces each sample's representation and allows the network to capture inter-sample dependencies within each episode. In parallel, \cite{santoro2016} augmented neural networks with an external memory module in which a recurrent controller, such as an LSTM, interacts with a differentiable memory matrix that stores representations of previously encountered examples. In this manner, the controller merges slow gradient-based learning with a fast memory-based adaptation. Other advancements include finding a model initialization that can be quickly fine-tuned to new tasks with only a few gradient steps and examples \cite{finn2017}, using meta-learners to generate fast-adapting parameters for few-shot tasks \cite{mann2017}, extending variational autoencoder architectures to operate on entire datasets instead of individual data points \cite{hewitt2018}, and replacing fixed distance functions with a learnable comparison module \cite{sung2018}.

In addition to the methods mentioned above, Prototypical Networks (ProtoNets) are extremely efficient and conceptually simple. The core idea behind ProtoNets is to compute the mean feature representation for each class, called the class prototype, and classify each new image by assigning it to a class whose prototype has the smallest Euclidean distance to the image \cite{snell2017}. Subsequent research has extended ProtoNets beyond the classification problem in the domain of Few-Shot Semantic Segmentation, where prototypes formed from annotated support masks guide pixel-level decisions in query images \hl{\cite{dong2018}, \cite{liu2020}, \cite{shaban2017},} as well as integrate task guidance \cite{rakelly2018}, \cite{wu2020} and iterative attention mechanisms and refinement \cite{zhang2019canet}, \cite{zhang2018sgone}.

Research on ProtoNets and other FSL methods has made significant progress, leading to the development and exploration of new, advanced techniques for Few-Shot Classification. A notable detail that can be noticed in the aforementioned studies is that the encoder architecture often corresponds to the overall functionality of the system, thereby exhibiting a form of structural parallelism. For example, \cite{koch2015}, which aims to determine whether two images are "same" or "different", employs an encoder composed of two identical CNNs sharing the same weights. Although precisely identifying the best encoder for a given methodology or problem remains nearly impossible owing to the black-box nature of Deep Neural Networks (DNNs), the observed parallelism between architecture and functionality offers an encouraging path for exploring new encoder designs. Given that Prototypical Networks use a simple convolutional block as the encoder, ProtoNets are ideal candidates for investigating improved encoder architectures \cite{snell2017}.

Despite the rapid evolution of Few-Shot Learning methodologies, little attention has been devoted to the architectural expressiveness of encoders. Existing approaches typically employ shallow, monolithic, or very deep, heavy encoder backbones, whose limited receptive diversity constrains the richness of the resulting feature space. This gap suggests that improvements in encoder design could yield substantial gains in representational capacity without altering the overarching meta-learning framework. Motivated by this observation, the next section introduces the Sum12 Multi-Scale encoder. This architecture leverages Multi-Scale feature extraction and integrates it into Prototypical Networks while maintaining simplicity and efficiency, yielding a lightweight CNN.

\section{Theoretical Background}\label{sec:theoretical_background}

Neural networks and artificial intelligence applications often seek to approximate biological information processing. For example, humans can recognize the same object across vastly different viewing distances and distinguish it from other objects by relying on several distinctive features that are already stored in memory. Consequently, representing visual information across multiple spatial scales is a fundamental property of natural intelligent vision systems. This was formalized scientifically in \hl{1994} with Tony Lindeberg's Scale-Space Theory \cite{lindeberg1994}, which established that any image representation consistent with the principles of linearity, shift invariance, and semigroup properties must be generated through convolution with a Gaussian kernel:

\begin{equation}
\label{eq:lindeberg_eq}
L(x, y; \sigma) = G(x, y; \sigma) * f(x, y)
\end{equation}
where $f(x, y)$ is the original image, $G(x, y; \sigma)$ is the Gaussian kernel with variance $\sigma^2$, and $*$ denotes convolution. Increasing $\sigma$ corresponds to observing structures at progressively coarser scales while 
avoiding spurious artifacts \cite{lindeberg1994}, \hl{\cite{lindeberg1998}}.

\hl{The proposed Sum12 encoder is only loosely inspired by this general idea of representing visual information at multiple spatial extents. It should not be interpreted as a discrete approximation of Gaussian scale-space. Sum12 branches are finite, learned, nonlinear transformations rather than Gaussian smoothing operators, and they are not constrained to satisfy the semigroup or non-creation-of-new-extrema properties of Gaussian scale-space. Likewise, the branch identifier does not correspond to the Gaussian scale parameter $\sigma$ or to a variance $\sigma^2$. The relationship between the two frameworks is therefore limited to this broad conceptual motivation.}

\begin{table}[t]
\caption{\textbf{Effective receptive field (RF) of each branch stack. Stride $=1$, no dilation. For a stack of kernels $\{k_i\}$, $\mathrm{RF} = 1 + \sum_i (k_i - 1)$.}}
\label{tab:rf_per_branch}
\setlength{\tabcolsep}{3pt}
\begin{tabular}{|p{35pt}|p{70pt}|p{75pt}|p{55pt}|}
\hline
Branch&
Kernel stack&
$\sum (k_i - 1)$&
RF \\
\hline
B1&
$3 + 3 + 3 + 3$&
$2 + 2 + 2 + 2 = 8$&
$1 + 8 = 9$ \\
B2&
$6 + 3 + 3$&
$5 + 2 + 2 = 9$&
$1 + 9 = 10$ \\
B3&
$3 + 6 + 3$&
$2 + 5 + 2 = 9$&
$1 + 9 = 10$ \\
B4&
$3 + 3 + 6$&
$2 + 2 + 5 = 9$&
$1 + 9 = 10$ \\
B5&
$9 + 3$&
$8 + 2 = 10$&
$1 + 10 = 11$ \\
B6&
$3 + 9$&
$2 + 8 = 10$&
$1 + 10 = 11$ \\
B7&
$6 + 6$&
$5 + 5 = 10$&
$1 + 10 = 11$ \\
\hline
\end{tabular}
\end{table}

\hl{With respect to CNNs, stacking spatial convolutional layers progressively enlarges the theoretical receptive field of units in deeper layers \cite{simonyan2014very}, \cite{luo2016}. In such networks, this receptive-field expansion is an indirect consequence of layer stacking rather than an explicitly defined Gaussian scale parameter. Table~\ref{tab:rf_per_branch} reports the theoretical receptive field associated with each Sum12 branch stack. These values describe the spatial extent of the corresponding convolutional paths, but they should not be interpreted as discrete Gaussian scales. Branches with the same theoretical receptive-field size but different kernel orderings remain distinct learned nonlinear transformations.

One way to gain more control over a CNN is to introduce a Multi-Scale design that expands convolutional kernel sizes laterally, rather than increasing depth, through multiple parallel branches operating over different spatial extents. This approach provides a structured mechanism for extracting and aggregating information from multiple receptive-field configurations, which is particularly relevant to the low-data regimes considered in Few-Shot Learning (FSL).

The lateral expansion of convolutional encoders has previously been explored in the Inception family of architectures, starting with the Inception module \cite{szegedy2015googlenet}, as well as in mixed-kernel and adaptive-kernel convolutional designs \cite{tan2019mixconv}, \cite{tek2021adaptive}, and in MSENet \cite{li2023mse}, a lightweight network proposed for fry counting in aquaculture.

The Inception module consists of three convolutional branches with kernel sizes of $1{\times}1$, $3{\times}3$, and $5{\times}5$, along with a max-pooling branch of size $3{\times}3$. Pointwise ($1{\times}1$) convolutions precede the $3{\times}3$ and $5{\times}5$ convolutions and follow the max-pooling operation to reduce dimensionality and improve computational efficiency~\cite{hua2018pointwise}.

MSENet is a three-column convolutional network designed to transform a grayscale fry image into a single-channel density map. In MSENet, the first column starts with a $9 \times 9$ convolution and continues with several $7 \times 7$ blocks, the second starts with a $7 \times 7$ convolution and continues with $5 \times 5$ blocks, and the third starts with a $5 \times 5$ convolution and continues with $3 \times 3$ blocks. Pooling layers are inserted after the first two blocks~\cite{li2023mse}.

MixConv~\cite{tan2019mixconv} implements branching differently. Instead of constructing parallel branches in the classical way, it partitions the input channels into groups and applies depthwise convolutions with different kernel sizes to the different groups. The processed channel groups are then recombined along the channel dimension.

Adaptive-kernel convolutional designs provide an alternative way of implementing multi-scale processing within a single layer. Tek et al.~\cite{tek2021adaptive} introduced filter-wise adaptive receptive fields within a convolutional layer by modulating each filter with a trainable Gaussian envelope function. The Gaussian-envelope formulation in \cite{tek2021adaptive} builds on Lindeberg’s generalized Gaussian scale-space framework \cite{lindeberg2011}.  This allows different filters in the same layer to learn different effective spatial extents, providing an implicit intra-layer form of multi-scale processing rather than an explicit multi-branch lateral expansion within the same filter.}

\hl{Motivated by these observations, we propose a Multi-Scale branching encoder whose central design constraint is that the kernel sizes within each branch sum to 12.}

\section{Sum12 Multi-Scale Branching Encoder Architecture}\label{sec:architecture}

The Sum12 Multi-Scale encoder begins with a base block consisting of two consecutive convolutional layers with $3{\times}3$ kernels, each followed by batch normalization and ReLU activation. This base block produces a shared feature map, which is then distributed to seven parallel convolutional branches. Each branch is composed of filters with kernel sizes of $3{\times}3$, $6{\times}6$, or $9{\times}9$. The depth of each branch, which can include two, three, or four convolutional layers, is chosen such that the sum of the kernel sizes within a branch equals 12. Specifically, one branch contains four $3{\times}3$ convolutions, one branch contains two $6{\times}6$ convolutions, two branches each combine one $3{\times}3$ and one $9{\times}9$ convolution, and the remaining three branches each combine two $3{\times}3$ convolutions and one $6{\times}6$ convolution. Together, these configurations exhaust the ordered additive arrangements of 3, 6, and 9 that sum to 12.

We impose the rule that the sum of the kernel sizes within each branch must equal 12, using kernels drawn from the set $\{3,6,9\}$. This constraint provides a compact and controlled receptive-field budget while allowing several ordered branch configurations. Other target sums are possible, such as 6, 9, 15, and 18; however, as shown in Section~\ref{sec:varying_sum_ablation}, increasing the target sum also increases the model weight. 

\begin{figure}[!htbp]
  \centering
  \includegraphics[width=0.78\linewidth]{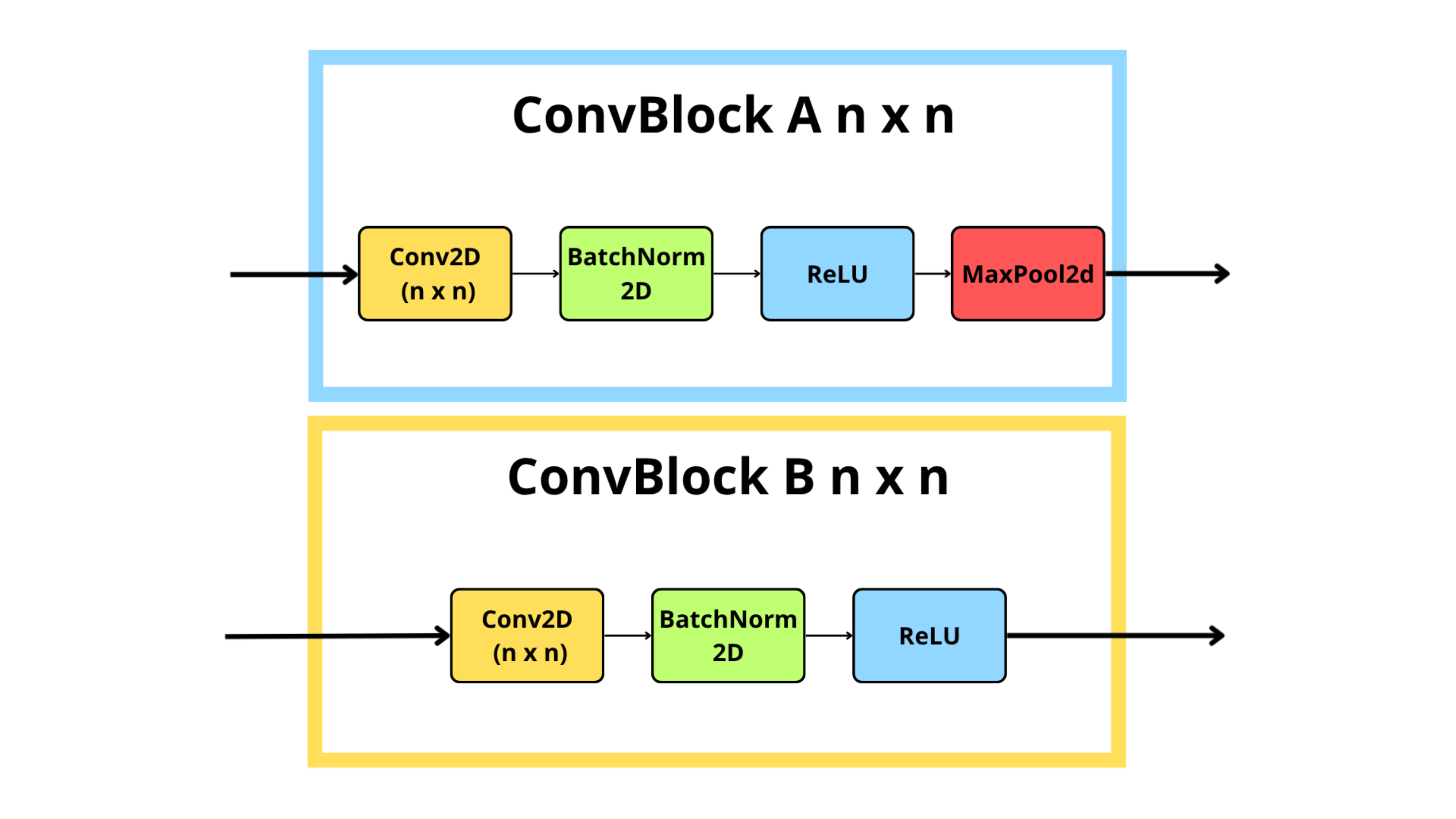}
  \caption{Convolution blocks used in the Sum12 Multi-Scale encoder. Convolution Block A is used for the double convolution before branching; Convolution Block B is used in the branches.}
  \label{fig:convBlocks}
\end{figure}

There is also a distinction between the two types of convolutional blocks used in the network: those applied to the input image and those employed within the branches. The former consists of a convolution $n{\times}n$, batch normalization, a ReLU activation, and a 2D max pooling layer. In contrast, the convolutional blocks used in the branches omit the pooling operation. This structural difference is illustrated in Figure~\ref{fig:convBlocks}.

Because larger kernels reduce the spatial extent of the output feature map, the results of the seven branches are center-cropped to a common spatial size $(\hat H, \hat W)$, denoted $\tilde F_k = \mathcal{C}_k(F_k)$. After alignment, the final feature representation is obtained through \hl{an element-wise arithmetic mean} as follows:
\begin{equation}
\label{Fsum_eq}
F_{\text{sum}} = \frac{1}{B}\sum_{k=1}^{B} \tilde F_k 
\end{equation}

where $B$ denotes the number of branches, which in our case equals $7$. In FSL experiments following the Prototypical Network methodology, the resulting embedding $\Phi(x) = F_{\text{sum}}$ is then passed to the metric stage, where the class prototypes are calculated as the mean of the support embeddings. The query samples are classified according to their Euclidean distance from each prototype.

\begin{figure*}[t]
  \centering
  \includegraphics[width=0.95\textwidth]{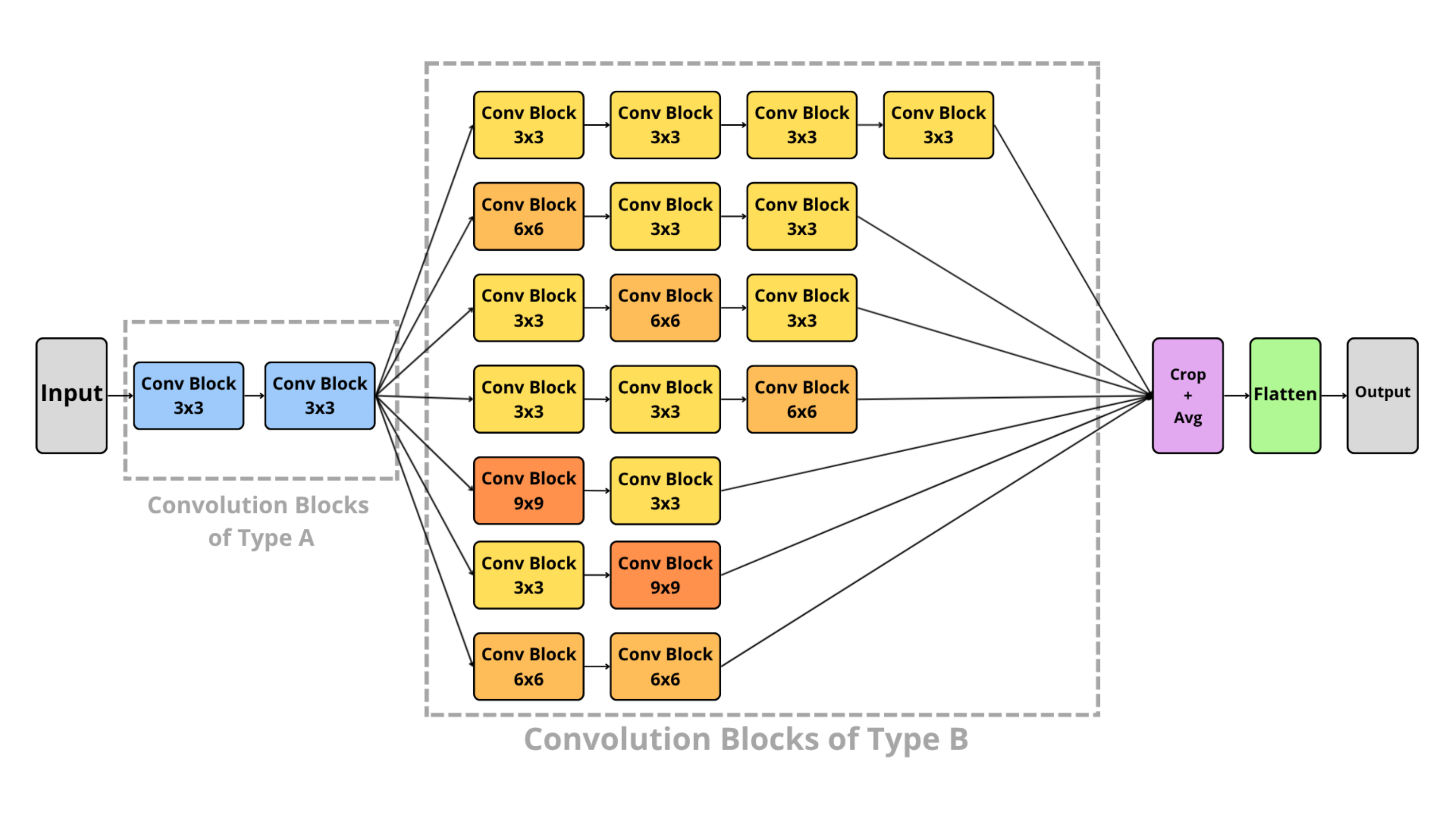}
  \caption{Proposed \emph{Sum12} encoder: seven parallel convolutional branches ($3{\times}3$, $6{\times}6$, $9{\times}9$); outputs are \hl{averaged} to form the embedding.}
  \label{fig:sum12_arch}
\end{figure*}

\hl{The proposed Sum12 encoder is shown in Figure~\ref{fig:sum12_arch}. The differences between the Sum12 Multi-Scale branching encoder and the other multi-scale architectures are as follows.

Inception and MSENet construct explicit parallel convolutional paths that process the same input using different receptive-field configurations, and concatenate the outputs of these paths. The Inception Module employs relatively shallow branches containing different kernel sizes, pointwise convolutions, and a separate pooling branch, whereas MSENet employs deeper convolutional columns. MixConv partitions the input channels into groups and applies depthwise convolutions with different kernel sizes to the different groups, which are subsequently recombined along the channel dimension.

In contrast, the Sum12 design is based on the idea that each ordered sequence of heterogeneous-kernel convolutions defines a distinct transformation. Sum12 therefore represents different kernel orderings as separate deep convolutional branches, distinguishing it from the other architectures considered here.

The final fusion operation is also different. Inception, MixConv, and MSENet all combine their respective outputs through channel-wise concatenation. By contrast, Sum12 computes an element-wise arithmetic mean of its spatially aligned branch outputs. This operation represents fixed uniform weighting, unlike the adaptive weighted combination used by Adaptive Kernel Convolution.

Among the architectures discussed above, MSENet may appear to be the most structurally similar to Sum12 because its columns are deep and contain several convolutional operations. Adaptive Kernel Convolution, on the other hand, shares the closest conceptual motivation with Sum12, as it also draws on Lindeberg's theory.

Unlike Sum12, which was designed as a FSL encoder, MSENet was developed for fish-fry counting and therefore designed to generate a spatial density map in which each location represents an estimated fry density. Furthermore, MSENet incorporates residual channel attention. Each MSENet column ends with a pre-activation residual block containing two $3{\times}3$ convolutions and a Squeeze-and-Excitation (SE) mechanism.

The conceptual difference between Sum12 and Adaptive Kernel Convolution lies in how the two architectures incorporate ideas from Lindeberg's theory. In~\cite{tek2021adaptive}, Gaussian functions are introduced explicitly as envelope functions. Sum12 does not directly implement Gaussian functions or the formal properties of Gaussian scale-space. It should instead be understood as being inspired only to a degree by multi-scale and scale-space principles rather than as a direct application of Lindeberg's theory in CNNs.

To summarize, the features that distinguish Sum12 from the above-mentioned architectures are the joint use of order-sensitive heterogeneous-kernel convolutional branches, a fixed cumulative kernel-size sum of 12 within each branch, and parameter-free element-wise mean fusion in a lightweight encoder designed for few-shot classification.}

A comparative analysis between the Inception module and the Sum12 Multi-Scale branching encoder, including the effect of the max-pooling branch on accuracy, is presented in Section~\ref{sec:Inception_ablation}.
Another difference is the use of concatenation in the Inception module. Concatenation combines features from different branches, preserving individual identities while increasing the number of channels because each branch contributes its own set of features. This is useful for Inception, a building block of larger networks, because it allows subsequent layers to learn combinations of features through later convolutions.

\hl{On the other hand, summation and averaging, as described in \mbox{\eqref{Fsum_eq}}, keep the number of channels constant, so the output has the same depth as the single-branch output. With respect to FSL, summation and averaging also serve as a form of regularization, reducing inter-branch variance and stabilizing learned features while encouraging the network to learn shared patterns across scales rather than treating them as separate.}

\section{Experimental Setup}\label{sec:experimental_setup}

This section presents the experimental setup used to evaluate the Sum12 Multi-Scale branching encoder. In the first subsection, we describe the four Few-Shot Learning datasets used to test the encoder within the Prototypical Network classification framework. In the second subsection, we introduce the MedMNIST 2D dataset and outline the experiments conducted in a conventional machine learning setting.

For all the experiments, we performed three independent training runs and reported their averages. For all datasets except BreastMNIST, the accuracy values exhibited low variance across the trained models. The high variance observed in the BreastMNIST dataset is explained in Section~\ref{sec:medmnist2d_results}.

\subsection{Few-Shot Learning Benchmarks}
We evaluated the effectiveness of the proposed Sum12 Multi-Scale encoder by conducting few-shot classification experiments using Omniglot, CIFAR-FS, FC100, and miniImageNet. We trained both the baseline Prototypical Network encoder and the Sum12 Multi-Scale encoder under identical episodic meta-learning conditions, using the same episode sampling strategy, optimizer, learning rate, and number of training episodes. This training regime ensures that we can attribute any performance differences solely to the encoder architecture rather than to training variations.

The encoder was trained for 10,000 episodic iterations following the protocol introduced in \cite{snell2017}. Below, we report the average classification accuracy obtained in 1,000 test episodes. The experiments were conducted in 5-Way and 20-Way settings, with 1-Shot and 5-Shot support configurations. For Omniglot, we also studied the evolution of accuracy across training episodes and reported the relative performance gains across training setups. Classification was performed using the standard prototype-based inference of Prototypical Networks \cite{snell2017}, which averages the support embeddings for each class to form a class prototype. Each query example was then classified based on the closest prototype in the embedding space.

\subsection{MedMNIST 2D}

Given that the Inception module is a conventional non-FSL methodology \cite{szegedy2015googlenet}, we also aimed to evaluate whether the proposed encoder can be effectively adapted to non-FSL benchmarks. A benchmark of particular interest is the MedMNIST 2D collection introduced in \cite{medmnist}. MedMNIST 2D comprises 12 datasets of various 2D medical images, each of which is resized to 28 $\times$ 28 pixels. The corresponding classification tasks include eight multi-class classification problems, two binary classification tasks, one multi-label binary classification task, and one regression task. Regarding dataset sizes, the smallest subsets are BreastMNIST (780 samples), RetinaMNIST (1,600 samples), and PneumoniaMNIST (5,856 samples). All remaining datasets contain more than 10,000 samples, with four exceeding 100,000 samples. The largest dataset, TissueMNIST, consists of 236,386 microscopic images of kidney cortex tissue. Lastly, of the 12 MedMNIST datasets, three \hl{are based on} abdominal CT scans. These are named OrganAMNIST, OrganCMNIST, and OrganSMNIST, where the letters A, C, and S stand for Axial, Coronal, and Sagittal views of the abdominal CT scan.

The authors of \cite{medmnist} also provide benchmark performance figures for several existing architectures and AutoML methodologies. Specifically, they trained and evaluated ResNet-18 and ResNet-50, as well as three automated learning systems: Auto-sklearn, AutoKeras, and Google AutoML Vision. Among these methods, AutoKeras and Google AutoML Vision employ Neural Architecture Search (NAS), whereas Auto-sklearn relies on ensemble methods and hyperparameter optimization. By training and evaluating the Sum12 Encoder on these datasets, we enable a direct comparison with widely adopted baselines, test its ability to generalize beyond the Few-Shot Learning domain, and demonstrate its applicability in a broader biomedical context.

For MedMNIST 2D, we trained the Sum12 Multi-Scale encoder in a standard supervised classification setting rather than in a few-shot episodic regime. For each dataset, we used the official training/validation/test splits provided in \cite{medmnist}. We trained the model from scratch using the cross-entropy loss and the Adam optimizer with a learning rate of $(1\times10^{-3})$. This training was conducted for 100 epochs with full-batch shuffling and without early stopping, using StepLR learning-rate decay with a decay factor $\gamma = 0.1$ at epochs 50 and 75.
During testing, we passed model outputs through softmax (or sigmoid for binary and multi-label tasks) and evaluated predictions using the official MedMNIST Evaluator utility, which reports both accuracy (ACC) and area under the ROC curve (AUC). In compiling results, the obtained metrics were averaged over three independent training runs.

We tested the ResNet-18 and ResNet-50 architectures at input resolutions of \(28 \times 28\) and \(224 \times 224\), with the latter obtained by upscaling the original \(28 \times 28\) images.

\section{Results and Discussion}\label{sec:results_discussion}
\subsection{Few-Shot Learning Baseline Datasets}

\subsubsection{Omniglot}

The first dataset we used was Omniglot, which is used very often in FSL experiments~\cite{vinyals2016, santoro2016, snell2017, finn2017, sung2018}. This dataset consists of 1,623 handwritten character classes collected from 50 distinct letters, ranging from real scripts such as Latin and Cyrillic to fictional scripts. Each class contains only 20 examples, yielding a total of approximately 32,460 grayscale images. This structure makes Omniglot a dataset characterized by very few samples per class but a large number of classes overall. Each grayscale character image has a small resolution of 28$\times$28 pixels. 

For the evaluation of Sum12 on Omniglot, we trained both the original Prototypical Network encoder and the proposed Sum12 Multi-Scale branching encoder at a 28$\times$28-pixel resolution. To ensure a perfectly controlled and repeatable comparison, the Prototypical Network was retrained using the same random seed, optimization settings, and convolutional kernel configuration of the Sum12 encoder. Models were trained for 2000 to 10000 episodic iterations, with performance evaluated every 2000 episodes using 1000 test episodes. This evaluation protocol allows us to analyze not only the final performance but also the convergence behavior, highlighting the differences in how quickly each architecture saturates its maximum accuracy.

The training results confirm that both encoders achieve very high accuracies, reflecting the relative simplicity of Omniglot compared to natural image datasets. As presented in \mbox{Table~\ref{tab:omniglot_full}}, the Sum12 Multi-Scale Branching Encoder consistently outperforms the original encoder. This improved performance is also confirmed by the gain heatmap in Figure~\ref{fig:gainheatmap}. Among the four test configurations, the 20-Way 1-Shot scenario shows the \hl{largest} gain for the Sum12 Multi-Scale Branching Encoder compared with the basic encoder used in Prototypical Networks. These results lead us to hypothesize that, in the Omniglot dataset, the performance gain increases with the difficulty of the $n$-Way $k$-Shot setting as $n$ increases. To verify this hypothesis, we conducted an additional experiment in which both encoders were trained in 5-Way and 30-Way configurations with 1-shot and 5-shot support sets and then tested in 30-Way and 60-Way configurations with 1-shot and 5-shot support sets. The results are presented in \mbox{Table~\ref{tab:30way_table}}. The gain ranges from +0.88\% to +2.56\% in various 5-Shot setups and from +2.34\% to +5.72\% in 1-Shot learning setups.

The most recent methodologies reporting exact Omniglot accuracies, namely Relation Networks \cite{sung2018} from 2018 and Multilevel Second-Order (MlSo) Few-Shot Learning \cite{MlSo} from 2022, surpass the accuracy of the Sum12 Multi-Scale Branching Encoder. For 5-Way 1-Shot, 5-Way 5-Shot, 20-Way 1-Shot, and 20-Way 5-Shot, Relation Networks report $99.6 \pm 0.2\%$, $99.8 \pm 0.1\%$, $97.6 \pm 0.2\%$, and $99.1 \pm 0.1\%$, respectively, whereas MlSo FSL reports 99.9\%, 99.9\%, 98.7\%, and 99.7\%. Referring to \mbox{Table~\ref{tab:omniglot_full}} in Appendix~\ref{app:omniglot_extended_results}, the Sum12 Multi-Scale Branching Encoder achieves its maximal accuracies when trained at 60-Way and evaluated in 5-Way or 20-Way scenarios. Taking into account MlSo, which is the better performer of the two comparison methods, the differences between MlSo and the best average accuracies of Sum12 are 0.7\%, 0.1\%, 1.8\%, and 0.5\%.

The advantage of our encoder is that, with respect to \cite{sung2018}, the performance difference remains below $1\%$ in all four standard $N$-Way $K$-Shot setups. With respect to \cite{MlSo}, it remains below $1\%$ in all cases except the 20-Way 1-Shot setting, while maintaining a significantly lower computational cost and a much smaller model size. More recent studies, such as \cite{luo2023} and \cite{guan2025}, consider such small differences to be relatively minor because they indicate that the data set has been saturated and suggest that other aspects should be considered when comparing the models.

MlSo constructs second-order feature representations by computing covariance matrices from convolutional feature maps at multiple network stages, thereby capturing pairwise co-activation relationships beyond first-order averaging. The construction of these covariance descriptors requires matrix multiplications between feature activations, which can scale from $O(d^2)$ to $O(d^3)$ depending on the feature dimensionality and normalization step. These matrices are subsequently power-normalized to suppress activation burstiness and stabilize statistical magnitude before being used in metric comparison. Although these second-order descriptors improve discrimination, they introduce substantial computational and memory overhead, increasing both training and inference costs.

In contrast, our encoder avoids covariance computation entirely, relying instead on a purely first-order, Multi-Scale feature representation that preserves efficiency and compactness while still yielding competitive class separability in the prototype space. \hl{This keeps the model size low and supports} its role as a lightweight encoder.

\hl{
In \cite{liu2020prototype_supp}, the supplementary material accompanying \cite{liu2020prototype}, Liu et al. report classification results for Cosine-Similarity-Based Prototypical Networks (CSPN) and Bias-Diminishing CSPN (BD-CSPN) under 5-Way 1-Shot and 5-Way 5-Shot evaluation. The two methods are evaluated using ConvNet-64, ConvNet-128, ConvNet-256, ResNet-12, and WRN-28-10 backbones. With ResNet-12, CSPN obtains accuracies of 98.70\% and 99.72\% in the 1-Shot and 5-Shot settings, respectively, while BD-CSPN obtains 99.80\% and 99.77\%. With WRN-28-10, CSPN achieves 99.02\% and 99.82\%, whereas BD-CSPN achieves 99.08\% and 99.85\% in the corresponding settings. These results also show that increasing backbone capacity does not uniformly improve every configuration, as ResNet-12 obtains the highest 1-Shot BD-CSPN accuracy. Because CSPN and BD-CSPN employ different training and inference procedures from those used for Sum12, and because BD-CSPN additionally uses transductive query-set information, the following comparisons are intended to contextualize accuracy rather than constitute strictly controlled comparisons with the more contemporary methods.

As shown in Table~\ref{tab:omniglot_full}, when Sum12 is trained and evaluated under the same 5-Way configurations, it obtains accuracies of 97.53\% and 99.31\% in the 1-Shot and 5-Shot settings, respectively. In the 1-Shot setting, Sum12 exceeds CSPN with the ConvNet-64 and ConvNet-128 backbones but remains below the other CSPN configurations and all reported BD-CSPN configurations. Its 5-Shot accuracy is below all the values reported by Liu et al. The highest Sum12 accuracies are obtained after training for 10,000 episodes using higher-way tasks. Specifically, training with 60-Way 1-Shot episodes yields an accuracy of 99.19\% under 5-Way 1-Shot evaluation, while training with 60-Way 5-Shot episodes yields 99.80\% under 5-Way 5-Shot evaluation. The latter result exceeds all the reported ConvNet and ResNet-12 results and is lower only than the WRN-28-10 CSPN and BD-CSPN results of 99.82\% and 99.85\%, respectively.

Lastly, Yoon et al.~\cite{yoon2019tapnet} report accuracies of 98.07\% and 99.49\% for TapNet under 20-Way 1-Shot and 20-Way 5-Shot evaluation, respectively, after meta-training with 60-Way episodes. By comparison, Sum12 achieves 96.46\% and 99.04\% when trained and evaluated under the corresponding 20-Way configurations, and 96.82\% and 99.20\% when trained using 60-Way 1-Shot and 60-Way 5-Shot episodes and subsequently evaluated under the corresponding 20-Way settings. Consequently, the Sum12 results obtained after 60-Way training provide the closest protocol comparison with TapNet, although the methods, backbone architectures, and numbers of training episodes still differ. In particular, TapNet was trained for 100{,}000 episodes, whereas the Sum12 experiments used 10{,}000 episodes.
}

\begin{table*}
\caption{\textbf{Omniglot stress-test: ProtoNet vs.\ \emph{Sum12}. Gains are $(\text{Sum12} - \text{ProtoNet})$.}}
\label{tab:30way_table}
\setlength{\tabcolsep}{3pt}
\begin{tabular}{|p{55pt}|p{31pt}|p{31pt}|p{25pt}|p{31pt}|p{31pt}|p{25pt}|}
\hline
Testing Config&
\multicolumn{3}{p{87pt}|}{Training A}&
\multicolumn{3}{p{87pt}|}{Training B} \\
\hline
&
Proto&
Sum12&
Gain&
Proto&
Sum12&
Gain \\
\hline
30-Way 1-Shot&
88.19\%&
\textbf{91.94\%}&
+3.75&
86.17\%&
\textbf{90.09\%}&
+3.92 \\
30-Way 5-Shot&
95.98\%&
\textbf{97.12\%}&
+1.14&
96.01\%&
\textbf{97.71\%}&
+1.70 \\
60-Way 1-Shot&
82.93\%&
\textbf{87.75\%}&
+4.82&
79.77\%&
\textbf{85.49\%}&
+5.72 \\
60-Way 5-Shot&
93.76\%&
\textbf{95.53\%}&
+1.77&
93.57\%&
\textbf{96.13\%}&
+2.56 \\
\hline
\end{tabular}
\end{table*}

\subsubsection{FC100}

The second benchmark is FC100, a dataset derived from CIFAR-100 and designed as a challenging few-shot learning dataset. Oreshkin et al.~\cite{oreshkin2018}  designed FC100 to reduce the information overlap between classes by splitting CIFAR-100 according to superclasses rather than individual classes. The dataset partitions 100 categories into 60 for training, 20 for validation, and 20 for testing, ensuring that evaluation occurs exclusively on novel classes. The images were resized to $84{\times}84$ pixels, and the models were trained and evaluated under the episodic meta-learning paradigm. In this way, the experimental setup promotes generalization to unseen tasks and discourages overfitting to low-level visual similarities.

\hl{Table~\ref{tab:fc100} reports the results obtained by ProtoNet and the proposed Sum12 Multi-Scale Branching Encoder across the different training and evaluation configurations, together with the corresponding accuracy differences between Sum12 and ProtoNet. Across the 16 comparisons, Sum12 outperforms ProtoNet in nine cases and underperforms it in seven. The differences range from a decrease of $1.24$ percentage points, observed when the models are trained using 5-Way 1-Shot episodes and evaluated in the 5-Way 5-Shot setting, to an improvement of $3.22$ percentage points, observed when they are trained using 20-Way 1-Shot episodes and evaluated in the 5-Way 5-Shot setting.}

\hl{When trained using the 5-Way 1-Shot configuration, Sum12 performs below ProtoNet in all four evaluation settings, with differences of $-0.81$, $-1.24$, $-0.60$, and $-0.61$ percentage points, respectively. Conversely, when trained using the 20-Way 1-Shot configuration, Sum12 outperforms ProtoNet in all four evaluation settings, with gains of $+1.41$, $+3.22$, $+1.06$, and $+2.79$ percentage points, respectively. The remaining 5-Way 5-Shot and 20-Way 5-Shot training configurations produce mixed results, thus the advantage of Sum12 on FC100 is not uniform across all experimental conditions and appears to depend on the episodic training configuration.}

\hl{
For comparison with deeper residual backbones, Lee et al.~ \cite{lee2019metaoptnet} report accuracies of 37.5\% and 52.5\% on FC100 under 5-Way 1-Shot and 5-Way 5-Shot evaluation, respectively, using Prototypical Networks with a ResNet-12 backbone. In comparison, when Sum12 is trained and evaluated under the corresponding 5-Way configurations, it obtains accuracies of 27.10\% and 42.89\% in the 1-Shot and 5-Shot settings, respectively. Furthermore, training using 20-Way 1-Shot episodes yields 29.17\% under 5-Way 1-Shot evaluation, while training using 20-Way 5-Shot episodes yields 46.09\% under 5-Way 5-Shot evaluation. Relative to Prototypical Networks with ResNet-12, these highest Sum12 results are lower by 8.33 and 6.41 percentage points in the 1-Shot and 5-Shot settings, respectively.

Other methods using ResNet-12 report even higher absolute accuracies. Lee et al.~\cite{lee2019metaoptnet} obtain 41.1\% and 55.5\% with MetaOptNet-SVM, while Tian et al.~\cite{tian2020rethinking} report 42.6\% and 59.1\% with RFS-simple and 44.6\% and 60.9\% with RFS-distill. For WRN-based models, Dhillon et al.~ \cite{dhillon2020baseline} report accuracies of 38.25\% and 57.19\% using inductive fine-tuning with a WRN-28-10 backbone. Their transductive fine-tuning method obtains 43.16\% and 57.57\%, although this configuration additionally uses information from the unlabeled query set during adaptation. These methods employ deeper backbones and different training, adaptation, and inference procedures from those used for Sum12; therefore, the above-mentioned comparisons are intended to contextualize the accuracy range in the dataset.
}

\hl{Although the improvements observed on FC100 are less consistent than those obtained on the other datasets, they can be contextualized within the broader range of results reported for early metric-based FSL methods. In the context of miniImageNet, the Matching Networks method with fine-tuning improves accuracy from a base range of $23.0\%$--$38.4\%$ to $41.2\%$--$46.6\%$ \cite{vinyals2016}. Likewise, under 5-Way 5-Shot evaluation, the Sum12 encoder achieves
$42.89\% \pm 0.48\%$ when trained using 5-Way 5-Shot episodes, or
$46.09\% \pm 0.48\%$ when trained using 20-Way 5-Shot episodes. These values are not intended as direct comparisons with FC100, since the datasets differ in their class structure and difficulty, but rather as contextual examples of the performance ranges reported for foundational metric-based FSL methods.}

\hl{Based on the FC100 results, although Sum12 does not consistently outperform the original ProtoNet encoder under every training configuration, it achieves higher accuracy in the majority of the reported comparisons and provides its most consistent improvements when trained using 20-Way 1-Shot episodes.}

\begin{table*}[t]
\caption{\textbf{FC100 ($84{\times}84$): ProtoNet vs.\ \emph{Sum12}. 
Best in each pair in \textbf{bold}. \hl{Gains are reported as 
$\emph{Sum12}-\mathrm{ProtoNet}$ in percentage points.} }}
\label{tab:fc100}
\begingroup
\rowcolors{1}{yellow}{yellow}

\setlength{\tabcolsep}{3pt}
\begin{tabular}{|p{55pt}|p{75pt}|p{75pt}|p{75pt}|p{75pt}|p{75pt}|}
\hline
Trained at &
Encoder &
5W-1S &
5W-5S &
20W-1S &
20W-5S \\
\hline

5W-1S &
ProtoNet &
\textbf{27.91\%} $\pm$ 0.42 &
\textbf{37.49\%} $\pm$ 0.45 &
\textbf{9.23\%} $\pm$ 0.13 &
\textbf{14.94\%} $\pm$ 0.14 \\

5W-1S &
\emph{Sum12} &
27.10\% $\pm$ 0.39 &
36.25\% $\pm$ 0.44 &
8.63\% $\pm$ 0.11 &
14.33\% $\pm$ 0.13 \\

&
\emph{Gain} &
$-0.81$ &
$-1.24$ &
$-0.60$ &
$-0.61$ \\
\hline

5W-5S &
ProtoNet &
28.30\% $\pm$ 0.39 &
41.33\% $\pm$ 0.45 &
\textbf{9.70\%} $\pm$ 0.13 &
17.89\% $\pm$ 0.15 \\

5W-5S &
\emph{Sum12} &
\textbf{28.35\%} $\pm$ 0.39 &
\textbf{42.89\%} $\pm$ 0.48 &
9.23\% $\pm$ 0.13 &
\textbf{19.36\%} $\pm$ 0.15 \\

&
\emph{Gain} &
$+0.05$ &
$+1.56$ &
$-0.47$ &
$+1.47$ \\
\hline

20W-1S &
ProtoNet &
27.76\% $\pm$ 0.41 &
36.67\% $\pm$ 0.44 &
9.23\% $\pm$ 0.12 &
14.64\% $\pm$ 0.14 \\

20W-1S &
\emph{Sum12} &
\textbf{29.17\%} $\pm$ 0.39 &
\textbf{39.89\%} $\pm$ 0.46 &
\textbf{10.29\%} $\pm$ 0.13 &
\textbf{17.43\%} $\pm$ 0.15 \\

&
\emph{Gain} &
$+1.41$ &
$+3.22$ &
$+1.06$ &
$+2.79$ \\
\hline

20W-5S &
ProtoNet &
\textbf{29.23\%} $\pm$ 0.40 &
45.27\% $\pm$ 0.48 &
\textbf{10.23\%} $\pm$ 0.13 &
20.74\% $\pm$ 0.17 \\

20W-5S &
\emph{Sum12} &
28.86\% $\pm$ 0.39 &
\textbf{46.09\%} $\pm$ 0.48 &
9.68\% $\pm$ 0.14 &
\textbf{21.48\%} $\pm$ 0.16 \\

&
\emph{Gain} &
$-0.37$ &
$+0.82$ &
$-0.55$ &
$+0.74$ \\
\hline

\end{tabular}
\endgroup
\end{table*}

\begin{table*}[t]
\caption{\textbf{CIFAR-FS ($32{\times}32$): ProtoNet vs.\ \emph{Sum12} across training/evaluation settings. Best in each pair in \textbf{bold}.}}
\label{tab:cifarfs}
\setlength{\tabcolsep}{3pt}
\begin{tabular}{|p{55pt}|p{55pt}|p{75pt}|p{75pt}|p{75pt}|p{75pt}|}
\hline
Trained at&
Encoder&
5W-1S&
5W-5S&
20W-1S&
20W-5S \\
\hline
5W-1S&
ProtoNet&
56.69\% $\pm$ 0.73&
72.02\% $\pm$ 0.60&
29.06\% $\pm$ 0.29&
45.47\% $\pm$ 0.30 \\
5W-1S&
\emph{Sum12}&
\textbf{62.82\%} $\pm$ 0.75&
\textbf{77.46\%} $\pm$ 0.57&
\textbf{35.42\%} $\pm$ 0.34&
\textbf{53.18\%} $\pm$ 0.30 \\
\hline

\rowcolor{yellow}
5W-5S&
ProtoNet&
56.47\% $\pm$ 0.76&
75.03\% $\pm$ 0.58&
30.02\% $\pm$ 0.32&
49.61\% $\pm$ 0.29 \\

\rowcolor{yellow}
5W-5S&
\emph{Sum12}&
\textbf{62.25\%} $\pm$ 0.70&
\textbf{80.20\%} $\pm$ 0.53&
\textbf{35.20\%} $\pm$ 0.31&
\textbf{57.47\%} $\pm$ 0.29 \\

\hline
20W-1S&
ProtoNet&
64.16\% $\pm$ 0.79&
78.39\% $\pm$ 0.57&
36.21\% $\pm$ 0.33&
54.37\% $\pm$ 0.32 \\
20W-1S&
\emph{Sum12}&
\textbf{71.93\%} $\pm$ 0.77&
\textbf{84.36\%} $\pm$ 0.47&
\textbf{45.81\%} $\pm$ 0.38&
\textbf{63.41\%} $\pm$ 0.32 \\
\hline
20W-5S&
ProtoNet&
58.84\% $\pm$ 0.76&
80.42\% $\pm$ 0.52&
32.63\% $\pm$ 0.34&
57.82\% $\pm$ 0.31 \\
20W-5S&
\emph{Sum12}&
\textbf{64.43\%} $\pm$ 0.74&
\textbf{84.05\%} $\pm$ 0.45&
\textbf{39.59\%} $\pm$ 0.38&
\textbf{64.73\%} $\pm$ 0.31 \\
\hline
\end{tabular}
\end{table*}

\subsubsection{CIFAR-FS}

The third dataset used for FSL testing was CIFAR-FS, another dataset derived from CIFAR-100, introduced by Bertinetto et al.~\cite{bertinetto2019}. CIFAR-FS uses a 64/16/20 class split for training, validation, and testing. Unlike FC100, it preserves the original class-level partitioning, thereby maintaining a higher visual overlap between classes. Each image is 32$\times$32 pixels, making the dataset computationally lightweight yet visually complex owing to its natural image variability. Consequently, this dataset serves as an intermediate benchmark as it is harder than Omniglot due to its natural textures and colors, yet less abstract than FC100 in terms of hierarchical dissimilarity. As an intermediate-complexity dataset, CIFAR-FS is used to compare the performance of the Sum12 Multi-Scale Branching Encoder with that of the Inception module.

The CIFAR-FS pipeline begins by loading the base CIFAR-100 dataset and splitting the classes using either the official JSON file or a deterministic fallback split, consisting of 64 training, 16 validation, and 20 test classes. The transforms consist of \texttt{ToTensor()} followed by normalization using the CIFAR-100 mean values (0.5071, 0.4865, 0.4409) and standard deviation values (0.2673, 0.2564, 0.2762). During training, \texttt{RandomHorizontalFlip} with $p=0.5$ is applied to support examples; no augmentation is applied to query examples or during validation and testing.

\begin{figure}[!htbp]
\centering
    \includegraphics[
        width=\linewidth,
        height=0.82\textheight,
        keepaspectratio
    ]{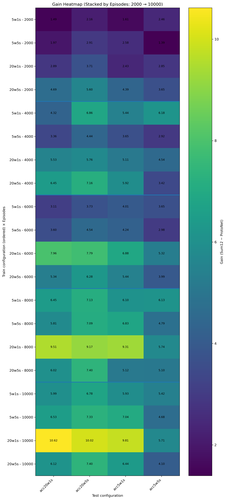}
    \caption{Gain heatmap on CIFAR-FS, stacked by training episodes.}
    \label{fig:cifarfs_gain_heatmap}
\end{figure}

\mbox{Table~\ref{tab:cifarfs}} reports results on CIFAR-FS across multiple cross-way generalization settings. Both encoders were trained under conventional 5-Way or 20-Way protocols and subsequently evaluated in 5-Way and 20-Way classification settings with 1-shot and 5-shot support sets. The baseline ProtoNet achieves competitive accuracy; however, its performance remains consistently below that of the Sum12 Multi-Scale Branching Encoder, particularly in the more demanding 20-Way evaluations. For example, when trained on 20-Way 1-Shot episodes, Sum12 improved upon ProtoNet by nearly 8 percentage points (71.9\% vs. 64.2\%) in the 5-Way 1-Shot test and by more than 9 percentage points (45.8\% vs. 36.2\%) in the 20-Way 1-Shot test. Similar gains are observed across other conditions, with the advantage of Multi-Scale feature extraction being most visible in high-way, low-shot regimes where finer discrimination is required. These findings \hl{show that the Sum12 Multi-Scale Branching Encoder consistently improves upon the ProtoNet baseline across the CIFAR-FS training and evaluation configurations considered here. An independently executed CIFAR-FS training-trajectory experiment, in which performance was evaluated at intermediate training episode counts, is reported in Table~\ref{tab:cifarfs_extended}. Because this training-trajectory experiment was executed separately from the experiments reported in Table~\ref{tab:cifarfs}, its final 10k results are not expected to reproduce exactly the values reported in Table~\ref{tab:cifarfs}. The corresponding gain heatmap is shown in Figure~\ref{fig:cifarfs_gain_heatmap}.
}

\hl{
Lee et al.~\cite{lee2019metaoptnet} report accuracies of 72.2\% and 83.5\% on CIFAR-FS under 5-Way 1-Shot and 5-Way 5-Shot evaluation, respectively, using Prototypical Networks with a ResNet-12 backbone. In comparison, when Sum12 is trained and evaluated under matching 5-Way configurations, it achieves 62.82\% in the 1-Shot setting and 80.20\% in the 5-Shot setting. However, higher-way training improves the Sum12 results substantially. Training under the 20-Way 1-Shot configuration yields 71.93\% under 5-Way 1-Shot evaluation and also produces the highest 5-Way 5-Shot result of 84.36\%. Relative to Prototypical Networks with ResNet-12, this Sum12 configuration is 0.27 percentage points lower in the 1-Shot setting but 0.86 percentage points higher in the 5-Shot setting.

Tian et al.~\cite{tian2020rethinking} report accuracies of 71.5\% and 86.0\% with RFS-simple and 73.9\% and 86.9\% with RFS-distill in the 1-Shot and 5-Shot settings, respectively. The RFS-trainval and RFS-distill-trainval variants achieve even higher results. For WRN-based models, Gidaris et al.~\cite{gidaris2019selfsupervision} report that Prototypical Networks with a WRN-28-10 backbone achieve 68.35\% and 81.79\% on CIFAR-FS. The best Sum12 results exceed these ProtoNet--WRN values numerically by 3.58 and 2.57 percentage points in the 1-Shot and 5-Shot settings, respectively. Still, they remain below state-of-the-art methods such as the cosine-classifier model with auxiliary rotation prediction, which achieves 73.62\% under 5-Way 1-Shot evaluation and 86.05\% under 5-Way 5-Shot evaluation.
}
\subsubsection{miniImageNet}

\begin{table*}
\caption{\textbf{Few-shot classification accuracies on miniImageNet. All accuracy results are averaged over 600 test episodes and are reported with 95\% confidence intervals.}}
\label{tab:miniImageNet_comparison}
\setlength{\tabcolsep}{3pt}
\begin{tabular}{|p{180pt}|p{50pt}|p{50pt}|p{75pt}|p{75pt}|}
\hline
Model&
Dist.&
Fine Tune&
5W-1S&
5W-5S \\
\hline
Baseline Nearest Neighbors \cite{snell2017}&
Cosine&
N&
28.86\% $\pm$ 0.54&
49.79\% $\pm$ 0.79 \\
Matching Networks \cite{vinyals2016}&
Cosine&
N&
43.40\% $\pm$ 0.78&
51.09\% $\pm$ 0.71 \\
Matching Networks FCE \cite{vinyals2016}&
Cosine&
N&
43.56\% $\pm$ 0.84&
55.31\% $\pm$ 0.73 \\
Meta-Learner LSTM \cite{ravi2017optimization}&
--&
N&
43.44\% $\pm$ 0.77&
60.60\% $\pm$ 0.71 \\
Prototypical Networks \cite{snell2017}&
Euclid.&
N&
49.42\% $\pm$ 0.78&
68.20\% $\pm$ 0.66 \\
Relation Network \cite{sung2018}&
--&
N&
50.44\% $\pm$ 0.82&
65.32\% $\pm$ 0.70 \\
Prototypical Networks (ours, reproduced)&
Euclid.&
N&
48.28\% $\pm$ 0.81&
66.12\% $\pm$ 0.64 \\
\emph{Sum12} Multi-Scale \hl{with adaptive pooling and SE-like gating} (ours)&
Euclid.&
N&
\textbf{52.81\% $\pm$ 0.85}&
\textbf{69.45\% $\pm$ 0.68} \\
\hline
\end{tabular}
\end{table*}

The fourth and final baseline benchmark used in the few-shot learning (FSL) experiments was miniImageNet. The miniImageNet dataset is a subset of the ImageNet dataset and consists of 100 classes, each containing 600 RGB images with a resolution of $84 \times 84 \times 3$. The dataset is split into 64 training classes, 16 validation classes, and 20 test classes, ensuring that evaluation is performed exclusively on previously unseen categories.

What distinguishes miniImageNet from the other baseline datasets is its pronounced sensitivity to small architectural and optimization changes. This sensitivity makes the reproducibility of previously reported results on miniImageNet particularly challenging, as minor variations in implementation details can lead to noticeable performance differences.

Because the goal of this study is to introduce a new compact encoder and compare it against established approaches, and because the original Prototypical Networks paper \cite{snell2017} does not provide exhaustive reproducibility details, we aimed to obtain results as close as possible to those reported in the original work under controlled and well-documented conditions.

We achieved comparable performance using the commonly adopted PKL-cached version of the miniImageNet dataset. To this end, we implemented a custom dataset class that loads images directly from pickle files and converts them using \texttt{Image.fromarray(img)}. The preprocessing pipeline consisted only of \texttt{ToTensor()}, and no normalization was applied in the referenced notebook. Consequently, the backbone received RGB images scaled to the \([0,1]\) range by \texttt{ToTensor()}, with all images stored at a resolution of \(84 \times 84 \times 3\). All experiments were conducted with CUDA, and the random seed was fixed at 42 to ensure reproducibility.

Furthermore, we used the crop and fusion methods in the miniImageNet experiments, which differed from those used for the other three datasets. Because cropping and averaging discard border information from branches with smaller kernels, we instead applied adaptive pooling combined with SE-like gating. \hl{Thus, the Sum12 Multi-Scale encoder used here specifically combines adaptive pooling and SE-like gating.} This approach was adopted as the first step towards exploring various crop and fusion strategies. A complete ablation study of crop and fusion strategies is presented in Subsection~\ref{sec:crop_fusion_ablations}.

We trained the model for up to 200 meta-epochs, each consisting of 400 episodes, although early stopping was triggered before reaching the maximum number of epochs in all cases. Following the evaluation protocol of \cite{snell2017}, we used two configurations: for the 1-shot scenario, we trained on 30-Way 1-Shot tasks and evaluated 5-Way 1-Shot tasks; for the 5-shot scenario, we trained on 20-Way 5-Shot tasks and evaluated 5-Way 5-Shot tasks. Under these settings, training the Prototypical Network encoder required 148 and 163 meta-epochs for the 1-shot and 5-shot cases, respectively, whereas training the proposed Sum12 Multi-Scale Branching Encoder required only 78 and 88 meta-epochs. In all experiments, we evaluated the model over 600 test episodes.

Using this setup, we obtained the accuracies of $48.28 \pm 0.81\%$ for the 1-shot case and $66.12 \pm 0.64\%$ for the 5-shot case when training the Prototypical Network, compared to the reported $49.42 \pm 0.78\%$ and $68.20 \pm 0.66\%$ in \cite{snell2017}. Under the same conditions, replacing the Prototypical Network encoder with the proposed Sum12 Multi-Scale Branching Encoder yielded accuracies of $52.81 \pm 0.85\%$ in the 1-shot scenario and $69.45 \pm 0.68\%$ in the 5-shot scenario. These results are higher than those reported for both Relation Networks and Prototypical Networks in the same 5-way miniImageNet settings. A quantitative comparison is presented in \mbox{Table~\ref{tab:miniImageNet_comparison}}.

\hl{Since the original random seeds and the exact determinism settings used in \cite{snell2017} are not available, our reproduced ProtoNet results are lower than the originally reported values by 1.14 percentage points in the 5-Way 1-Shot setting and by 2.08 percentage points in the 5-Way 5-Shot setting. However, the reproduced ProtoNet and Sum12 models were trained and evaluated using the same random seed and controlled deterministic settings. Therefore, Table~\ref{tab:miniImageNet_comparison} emphasizes the relative improvement obtained by replacing the reproduced ProtoNet encoder with Sum12. Under these controlled conditions, Sum12 improves upon the reproduced ProtoNet baseline by $52.81\%-48.28\%=4.53$ percentage points in the 5-Way 1-Shot setting and by $69.45\%-66.12\%=3.33$ percentage points in the 5-Way 5-Shot setting. }

\hl{To place these controlled improvements within the broader deep-backbone literature, Lee et al.~\cite{lee2019metaoptnet} report accuracies of 59.25\% and 75.60\% on miniImageNet under 5-Way 1-Shot and 5-Way 5-Shot evaluation, respectively, using Prototypical Networks with a ResNet-12 backbone. Relative to these results, Sum12 is lower by 6.44 percentage points in the 1-Shot setting and by 6.15 percentage points in the 5-Shot setting. Although both methods are based on the Prototypical Networks framework, the comparison is not strictly controlled because the backbone architectures, regularization methods, training procedures, and episodic configurations differ. In particular, Lee et al. use 5-Way meta-training and evaluation for their ResNet-12 ProtoNet experiments, whereas the Sum12 experiments follow the higher-way training protocol used in the original Prototypical Networks study.}

\hl{ Stronger results are reported when ResNet-12 is combined with alternative representation-learning and classification procedures. Tian et al.~\cite{tian2020rethinking} obtain 62.02\% and 79.64\% with RFS-simple and 64.82\% and 82.14\% with RFS-distill in the 1-Shot and 5-Shot settings, respectively. For WRN-based models, Gidaris et al.~\cite{gidaris2019selfsupervision} report accuracies of 55.85\% and 68.72\% using Prototypical Networks with a WRN-28-10 backbone. Sum12 is numerically 3.04 percentage points lower in the 1-Shot setting and 0.73 percentage points higher in the 5-Shot setting than this particular ProtoNet--WRN implementation. However, the stronger cosine-classifier model with auxiliary rotation prediction achieves 62.93\% and 79.87\%, remaining above Sum12 in both settings. }

\subsection{Few-Shot Learning Ablations}\label{sec:ablations}

Having established \hl{the performance of the Sum12 Multi-Scale Branching Encoder} on the four baseline datasets described above, we now consider several ablations and variations of the architecture.

\subsubsection{Crop and Fusion Strategies}\label{sec:crop_fusion_ablations}

First, as mentioned when discussing the complexity of the miniImageNet dataset, the center-crop, collapse-type, and averaging operations used in the \hl{initial Sum12 implementation} discard border information from branches with smaller kernels. To determine whether alternative crop or fusion methods were more effective than the initial approach, we conducted several experiments. These experiments were divided into two categories: testing several crop strategies with a fixed fusion method, namely averaging, and testing several fusion strategies with a fixed crop method.

Although \cite{snell2017} already established the advantage of Euclidean distance, we also performed a cross-verification between Euclidean and cosine distances during the crop-strategy tests, as shown in \mbox{Table~\ref{tab:omniglot_crop_strategy}} in Appendix~\ref{app:sum12_ablations}. This verification was conducted on the Omniglot data set because the crop strategy had changed. The results indicated that the Euclidean distance consistently outperformed the cosine distance, regardless of the crop strategy, by a large margin, sometimes around 20\%. Given these results, we did not continue testing with cosine distance in the remaining ablations.

We tested the fusion strategies on Omniglot and CIFAR-FS, and the results are reported in \mbox{Table~\ref{tab:omniglot_fusion_strategies}} and \mbox{Table~\ref{tab:cifarfs_fusion_strategies}} in Appendix~\ref{app:sum12_ablations}, respectively. We compared the original uniform averaging across branches with scalar per-branch weights, SE-like gating, weighted averaging, concatenation followed by a \(1\times1\) convolution, an attention mechanism, and a simple \(1\times1\) convolution.

A cross-verification of these two tables indicates that SE-like gating is the strongest fusion strategy in most cases. It ranks first or tied on seven of the 16 total Omniglot configurations, as shown in \mbox{Table~\ref{tab:omniglot_fusion_strategies}}, and in all one-shot and five-shot scenarios on CIFAR-FS, as shown in \mbox{Table~\ref{tab:cifarfs_fusion_strategies}}.

\subsubsection{Dilated Convolution Analysis}\label{sec:dilatation_analysis}

Another possible modification is to replace the \(6\times6\) and \(9\times9\) convolutions with dilated convolutions and to compare the two approaches in terms of classification accuracy, FLOPs, and model size.

\mbox{Table~\ref{tab:omniglot_dilatation}} and \mbox{Table~\ref{tab:cifarfs_dilatation}} present the accuracy comparison between the standard Sum12 encoder, which uses \(6\times6\) and \(9\times9\) convolutions, and the Sum12 variant based on dilated convolutions. \hl{The relative accuracy of the two variants is configuration-dependent rather than uniformly favoring either architecture. On Omniglot, the standard Sum12 encoder achieves higher accuracy in 13 of the 16 reported training--evaluation combinations, with a mean accuracy of approximately 97.59\% across all configurations compared with 97.41\% for the dilated variant. On CIFAR-FS, the standard encoder achieves higher accuracy in 12 of the 16 training--evaluation combinations and attains a mean accuracy of approximately 61.39\%, compared with 60.97\% for the dilated variant. Under the 5-Way 5-Shot training configuration, the dilated variant performs better in three of the four evaluation settings, whereas the standard Sum12 encoder performs better under 20-Way 1-Shot evaluation, achieving 35.20\% compared with 34.85\%. In this training regime, the largest gains of the dilated model are 1.14 and 1.22 percentage points under 5-Way 5-Shot and 20-Way 5-Shot evaluation, respectively.}

However, the dilated version is substantially lighter in terms of model size, requiring approximately \hl{a disk size of 38.5\% of} the original Sum12 encoder. The corresponding FLOPs and parameter counts are shown in \mbox{Table~\ref{tab:sum_12_flops}}. As expected, these architecture-level values are invariant to the episodic configuration, namely the \(N\)-way and \(K\)-shot settings. \hl{Therefore, using dilated convolutions substantially reduces the computational and parameter requirements, while producing configuration-dependent changes in classification accuracy rather than a consistent loss in performance.}

\begin{table*}
\centering
\caption{\textbf{Encoder FLOPs and parameter counts for the standard and dilated \emph{Sum12} variants, computed using the actual experimental input dimensions.}}
\label{tab:sum_12_flops}
\setlength{\tabcolsep}{3pt}
\begingroup
\rowcolors{1}{yellow}{yellow}

\begin{tabular}{|p{115pt}|p{80pt}|p{65pt}|}
\hline
Encoder Variant&
FLOPs&
Parameters \\
\hline
\emph{Sum12} Encoder (Omniglot)&
113,480,704&
1,884,672 \\
\emph{Sum12} Encoder (CIFAR-FS)&
146,726,272&
1,885,824 \\
\emph{Sum12} Dilated (Omniglot)&
42,241,920&
740,544 \\
\emph{Sum12} Dilated (CIFAR-FS)&
56,352,768&
741,696 \\
\hline
\end{tabular}
\endgroup
\end{table*}

\begin{table*}
\centering
\caption{\textbf{Omniglot average classification accuracy (\%) of the standard \emph{Sum12} encoder versus \emph{Sum12} with dilated convolutions.}}
\label{tab:omniglot_dilatation}
\setlength{\tabcolsep}{3pt}
\begin{tabular}{|p{55pt}|p{80pt}|p{55pt}|p{55pt}|p{55pt}|p{55pt}|}
\hline
Trained at&
Encoder&
5W-1S&
5W-5S&
20W-1S&
20W-5S \\
\hline
5W-1S&
\emph{Sum12}&
\textbf{97.53}&
\textbf{99.31}&
\textbf{93.01}&
\textbf{97.94} \\
5W-1S&
\emph{Sum12} Dilated&
97.25&
99.26&
92.75&
97.56 \\
\hline
5W-5S&
\emph{Sum12}&
\textbf{97.14}&
\textbf{99.53}&
\textbf{92.13}&
\textbf{98.14} \\
5W-5S&
\emph{Sum12} Dilated&
96.91&
99.29&
91.55&
98.00 \\
\hline
20W-1S&
\emph{Sum12}&
\textbf{98.83}&
99.67&
\textbf{96.46}&
\textbf{99.07} \\
20W-1S&
\emph{Sum12} Dilated&
98.61&
\textbf{99.72}&
96.15&
98.95 \\
\hline
20W-5S&
\emph{Sum12}&
98.36&
\textbf{99.73}&
\textbf{95.56}&
99.04 \\
20W-5S&
\emph{Sum12} Dilated&
\textbf{98.45}&
99.72&
95.36&
\textbf{99.08} \\
\hline
\end{tabular}
\end{table*}

\begin{table*}[t]
\caption{\textbf{CIFAR-FS classification accuracy (\%) of the standard \emph{Sum12} encoder versus \emph{Sum12} with dilated convolutions.}}
\label{tab:cifarfs_dilatation}
\centering
\setlength{\tabcolsep}{3pt}
\begin{tabular}{|p{55pt}|p{80pt}|p{75pt}|p{75pt}|p{75pt}|p{75pt}|}
\hline
Trained at&
Encoder&
5W-1S&
5W-5S&
20W-1S&
20W-5S \\
\hline
5W-1S&
\emph{Sum12}&
\textbf{62.82 $\pm$ 0.75}&
\textbf{77.46 $\pm$ 0.57}&
\textbf{35.42 $\pm$ 0.34}&
\textbf{53.18 $\pm$ 0.30} \\
5W-1S&
\emph{Sum12} Dilated&
62.52 $\pm$ 0.76&
76.84 $\pm$ 0.56&
34.28 $\pm$ 0.32&
51.68 $\pm$ 0.30 \\
\rowcolor{yellow}
5W-5S&
\emph{Sum12}&
62.25 $\pm$ 0.70 &
80.20 $\pm$ 0.53&
\textbf{35.20 $\pm$ 0.31}&
57.47 $\pm$ 0.29 \\
5W-5S&
\emph{Sum12} Dilated&
\textbf{62.49 $\pm$ 0.74}&
\textbf{81.34 $\pm$ 0.50}&
34.85 $\pm$ 0.33&
\textbf{58.69 $\pm$ 0.30} \\
\hline
20W-1S&
\emph{Sum12}&
\textbf{71.93 $\pm$ 0.77}&
\textbf{84.36 $\pm$ 0.47}&
\textbf{45.81 $\pm$ 0.38}&
\textbf{63.41 $\pm$ 0.32} \\
20W-1S&
\emph{Sum12} Dilated&
71.28 $\pm$ 0.78&
83.16 $\pm$ 0.50&
44.74 $\pm$ 0.38&
62.39 $\pm$ 0.33 \\
\hline
20W-5S&
\emph{Sum12}&
\textbf{64.43 $\pm$ 0.74}&
84.05 $\pm$ 0.45&
\textbf{39.59 $\pm$ 0.38}&
\textbf{64.73 $\pm$ 0.31} \\
20W-5S&
\emph{Sum12} Dilated&
64.19 $\pm$ 0.76&
\textbf{84.34 $\pm$ 0.48}&
38.10 $\pm$ 0.35&
64.61 $\pm$ 0.29 \\
\hline
\end{tabular}
\end{table*}

\subsubsection{Varying Sum of Encoder Branches}\label{sec:varying_sum_ablation}

The initial concept of branching the encoder after the first two convolutions into parallel paths, with kernel sizes summing to a fixed constant, was motivated by the hypothesis that the total receptive-field budget significantly influences representation quality. To test this hypothesis, the target sum was varied from 3 to 18, and the resulting architectures were evaluated under identical few-shot learning settings. The results are listed in \mbox{Table~\ref{tab:varying_sum_ablation}}.

\begin{figure}[!t]
  \centering
  \includegraphics[width=1\linewidth]{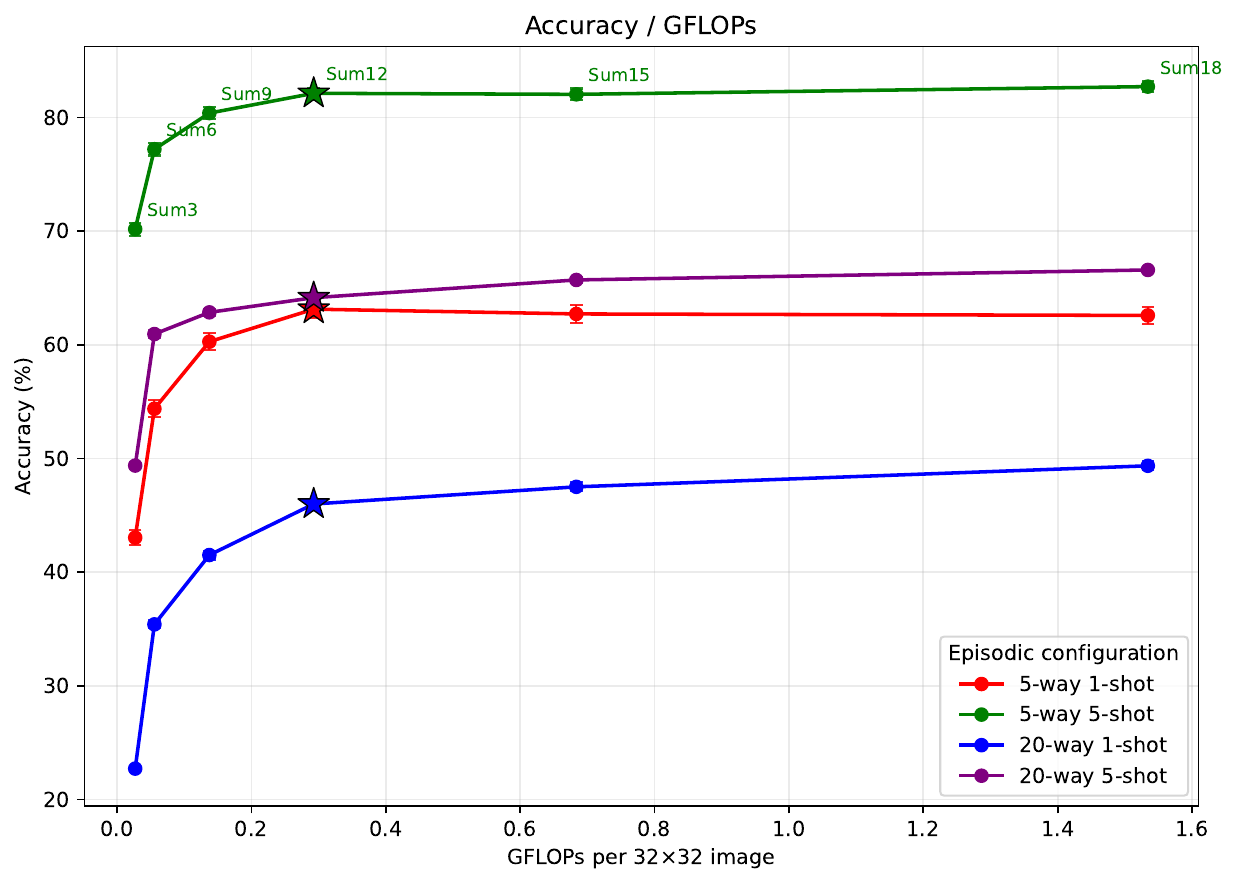}
  \caption{\hl{Accuracy versus GFLOPs for the Sum-variation architectures.}}
  \label{fig:sum_variations_accuracy_vs_GFLOPS}
\end{figure}

\begin{figure}[!t]
  \centering
  \includegraphics[width=1\linewidth]{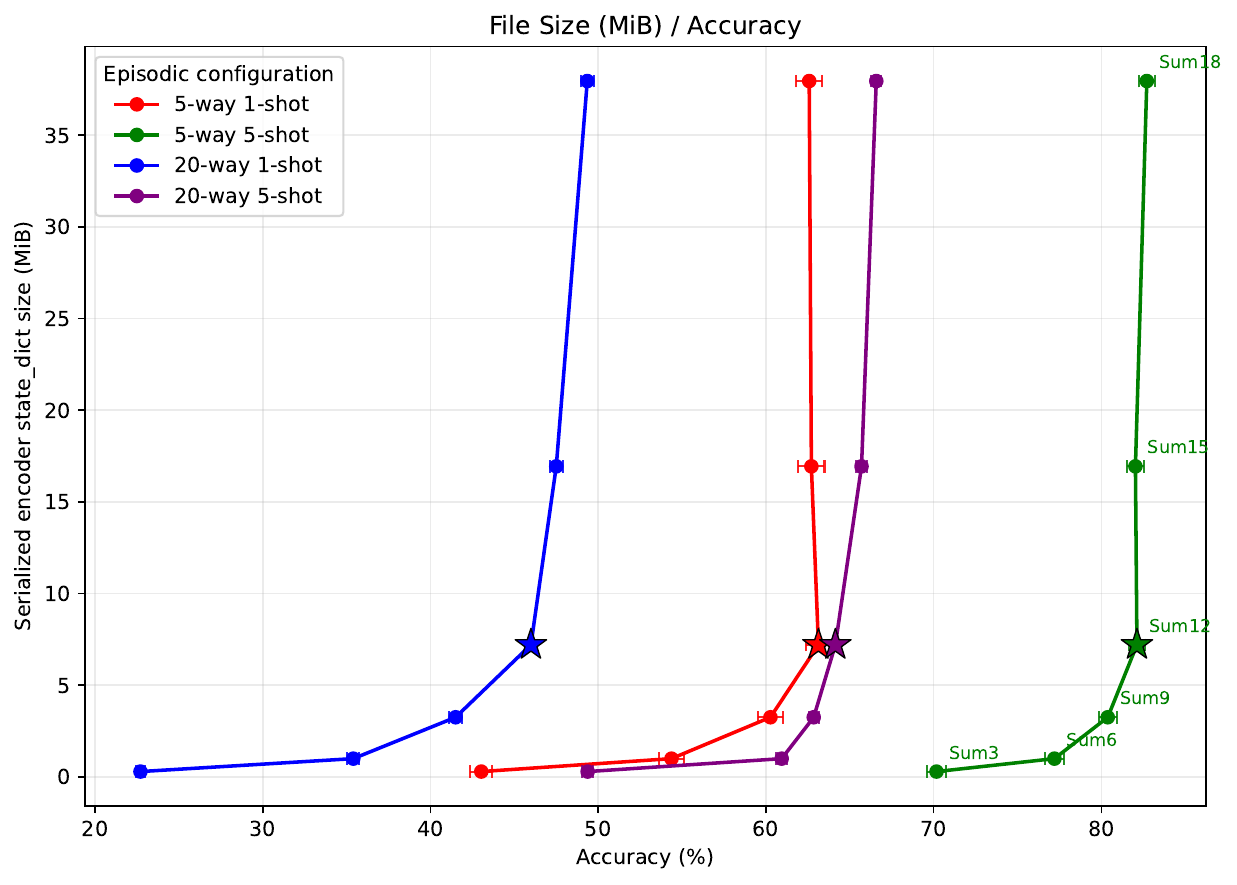}
  \caption{\hl{Serialized encoder size versus accuracy for the Sum-variation architectures.}}
  \label{fig:sum_variations_filesize_vs_accuracy}
\end{figure}

\hl{The reason why Sum12 remains the main architecture proposed in this
paper, although Sum15 and Sum18 yield better results in the majority of
cases, is explained as follows. Consider
Figure~\ref{fig:sum_variations_accuracy_vs_GFLOPS} and
Figure~\ref{fig:sum_variations_filesize_vs_accuracy}, which show the
relationship between accuracy and GFLOPs, and between serialized encoder
size and accuracy, respectively, for all the architectures from Sum3 to
Sum18 included in this ablation.

It can be seen from the graphs that the amount by which accuracy
increases generally becomes smaller as we move from one encoder to the
next. Thus, the accuracy increase from Sum3 to Sum6 is greater than the
increase from Sum6 to Sum9, and so on. At the same time, the resources
against which accuracy is compared, namely GFLOPs and the serialized
size of the saved model, continue to increase with each successive
architecture. This behavior is particularly visible in the transitions
from Sum12 to Sum15 and from Sum15 to Sum18.

In order to quantify the trade-off between the accuracy results reported in Table~\ref{tab:varying_sum_ablation} and architectural complexity, we selected the two resources considered above, GFLOPs and serialized encoder size, and calculated the marginal increase in resource consumption per percentage-point increase in accuracy between each pair
of consecutive architectures. Let $s_1=3$, $s_2=6$, $s_3=9$, $s_4=12$, $s_5=15$, and $s_6=18$ denote the ordered target sums. Let $A_{s_j}$ denote the accuracy of architecture Sum$s_j$, and let $R_{s_j}$ denote its resource consumption, where $R$ represents either GFLOPs or serialized encoder size. Being the transition of the first pair in the considered architectures, the Sum3-to-Sum6 transition was used as the reference marginal cost:}

\begin{equation}
\label{resource_cost_reference}
\hleq{
C_{\mathrm{ref}}^{(R)}
=
\frac{R_{s_2}-R_{s_1}}{A_{s_2}-A_{s_1}}
=
\frac{R_6-R_3}{A_6-A_3}
}
\end{equation}

\hl{We then used this reference value to normalize the same
relationship for each pair of consecutive architectures, according to:}

\begin{equation}
\label{resource_cost_normalized}
\hleq{
\widetilde{C}_{j}^{(R)}
=
\frac{
    \dfrac{R_{s_j}-R_{s_{j-1}}}
          {A_{s_j}-A_{s_{j-1}}}
}{
    C_{\mathrm{ref}}^{(R)}
},
\qquad j \in \{3,4,5,6\}
}
\end{equation}

\hl{This choice does not imply that the Sum3-to-Sum6 step has any privileged theoretical status, but it does provide a fixed baseline for comparing later transitions. The normalized marginal cost ratios are defined only when the accuracy differences in both the reference transition and the evaluated transition are nonzero.}

\begin{table*}[t]
\caption{\hl{\textbf{Normalized marginal file-size and GFLOPs cost ratios for the \emph{Sum} variations, expressed as multiples of the \emph{Sum3}-to-\emph{Sum6} transition baseline.}}}
\label{tab:sum_variations_resource_multiples}
\centering
\setlength{\tabcolsep}{4pt}

\begingroup
\rowcolors{1}{yellow}{yellow}

\resizebox{\textwidth}{!}{%
\begin{tabular}{|p{55pt}|
p{65pt}|p{65pt}|
p{65pt}|p{65pt}|
p{65pt}|p{65pt}|
p{65pt}|p{65pt}|}
\hline
Architecture&
\shortstack{5W-1S\\File size}&
\shortstack{5W-1S\\GFLOPs}&
\shortstack{5W-5S\\File size}&
\shortstack{5W-5S\\GFLOPs}&
\shortstack{20W-1S\\File size}&
\shortstack{20W-1S\\GFLOPs}&
\shortstack{20W-5S\\File size}&
\shortstack{20W-5S\\GFLOPs} \\
\hline

\emph{Sum6}&
1.0$\times$&
1.0$\times$&
1.0$\times$&
1.0$\times$&
1.0$\times$&
1.0$\times$&
1.0$\times$&
1.0$\times$ \\
\hline

\emph{Sum9}&
6.2$\times$&
5.5$\times$&
7.1$\times$&
6.3$\times$&
6.7$\times$&
6.0$\times$&
19.4$\times$&
17.3$\times$ \\
\hline

\emph{Sum12}&
22.2$\times$&
21.5$\times$&
22.6$\times$&
21.9$\times$&
15.8$\times$&
15.3$\times$&
50.2$\times$&
48.7$\times$ \\
\hline

\emph{Sum15}&
-372.4$\times$&
-368.9$\times$&
-1077.4$\times$&
-1067.2$\times$&
115.9$\times$&
114.8$\times$&
102.3$\times$&
101.3$\times$ \\
\hline

\emph{Sum18}&
-2598.2$\times$&
-2593.5$\times$&
307.9$\times$&
307.4$\times$&
204.3$\times$&
204.0$\times$&
391.6$\times$&
390.9$\times$ \\
\hline

\end{tabular}%
}
\endgroup
\end{table*}

\hl{In both resource measures, the magnitudes of the normalized marginal cost ratios generally become larger for the Sum15 and Sum18 architectures, indicating diminishing or occasionally negative accuracy gains despite increased resource consumption. When the accuracy difference between consecutive architectures approaches zero, this ratio can become very large; therefore, the normalized values should be interpreted as a descriptive indicator of diminishing returns rather than as a standalone criterion for architectural selection. In cases where the accuracy change is negative, indicating that the larger architecture achieves lower accuracy than the preceding one, the absolute value of the ratio reflects the magnitude of this unfavorable resource-to-accuracy trade-off.
As shown in Table~\ref{tab:sum_variations_resource_multiples}, the normalized ratios beyond Sum12 can become substantially larger than the reference transition, although Sum15 and Sum18 still provide meaningful accuracy improvements in some configurations.}

\hl{Accordingly, Sum12 is selected as the main architecture based on two practical criteria. First, it is the smallest evaluated variant at which the Accuracy--GFLOPs and Accuracy--Model-Size curves begin to exhibit diminishing returns. Second, the embedded study in Section~\ref{sec:section_stm32_deployment} shows that even the compressed Sum12 variant lies close to the practical resource limit of the investigated STM32 configuration.}

\subsubsection{Branch Count and Composition Ablation}\label{sec:branch_composition_ablation}

Because the Sum12 encoder is constructed by combining all seven possible configurations of kernel sizes drawn from $\{3, 6, 9\}$ whose sum equals 12, it is necessary to assess the contribution of each branch to the overall performance. To this end, we adopted a two-step evaluation strategy.

First, we evaluated the effectiveness of each branch by treating it as a standalone encoder and comparing its performance against the average accuracy of the corresponding configuration. CIFAR-FS was selected for this analysis because it is a more challenging dataset and therefore exhibits more pronounced performance differences. The results are reported in \mbox{Table~\ref{tab:cifarfs_branches_individually}} in Appendix~\ref{app:sum12_ablations}, where branches with contributions below the average are highlighted in italics. Based on this analysis, we constructed two modified Sum12 variants: \emph{Sum12 Reduced}, in which the less effective branches were removed, and \emph{Sum12 Duplicated}, in which duplicates of the more effective branches replaced them.

In the second step, both variants were evaluated using Omniglot and CIFAR-FS. The Omniglot results are reported in \mbox{Table~\ref{tab:omniglot_sum12_variants_duplicated_reduced}}, whereas the CIFAR-FS results are reported in \mbox{Table~\ref{tab:cifarfs_sum12_variants_duplicated_reduced}}. In Omniglot, the results were mixed, and the performance differences between the standard Sum12 encoder, Sum12 Reduced, and Sum12 Duplicated remained marginal. However, on CIFAR-FS, the performance variations across architectures and training configurations were more pronounced. Overall, the best results were achieved by the standard Sum12 encoder trained in the 20-Way 1-Shot setting when evaluated on the 5-Way \(K\)-Shot tasks. At the same time, Sum12 Duplicated yielded the highest accuracy for both 20-Way 1-Shot and 20-Way 5-Shot evaluations when trained under the corresponding configurations.

\begin{table*}[t]
\caption{\textbf{Comparison of \emph{Sum12} variants on Omniglot. Classification accuracy (\%).}}
\label{tab:omniglot_sum12_variants_duplicated_reduced}
\centering
\setlength{\tabcolsep}{3pt}
\begin{tabular}{|p{55pt}|p{90pt}|p{75pt}|p{75pt}|p{75pt}|p{75pt}|}
\hline
Trained at&
Encoder&
5W-1S&
5W-5S&
20W-1S&
20W-5S \\
\hline
5W-1S&
\emph{Sum12}&
\textbf{97.72 $\pm$ 0.26}&
99.26 $\pm$ 0.13&
93.38 $\pm$ 0.22&
97.77 $\pm$ 0.13 \\
5W-1S&
\emph{Sum12} Reduced&
\textbf{97.72 $\pm$ 0.26}&
\textbf{99.43 $\pm$ 0.11}&
93.32 $\pm$ 0.22&
\textbf{98.15 $\pm$ 0.10} \\
5W-1S&
\emph{Sum12} Duplicated&
97.67 $\pm$ 0.27&
99.34 $\pm$ 0.14&
\textbf{93.79 $\pm$ 0.22}&
98.09 $\pm$ 0.12 \\
\hline
5W-5S&
\emph{Sum12}&
\textbf{97.34 $\pm$ 0.27}&
\textbf{99.44 $\pm$ 0.10}&
\textbf{92.29 $\pm$ 0.23}&
\textbf{98.27 $\pm$ 0.09} \\
5W-5S&
\emph{Sum12} Reduced&
96.51 $\pm$ 0.30&
99.21 $\pm$ 0.12&
91.15 $\pm$ 0.24&
97.53 $\pm$ 0.12 \\
5W-5S&
\emph{Sum12} Duplicated&
96.94 $\pm$ 0.30&
99.37 $\pm$ 0.11&
91.84 $\pm$ 0.23&
97.98 $\pm$ 0.10 \\
\hline
20W-1S&
\emph{Sum12}&
\textbf{98.94 $\pm$ 0.19}&
99.78 $\pm$ 0.07&
\textbf{96.48 $\pm$ 0.17}&
\textbf{99.14 $\pm$ 0.07} \\
20W-1S&
\emph{Sum12} Reduced&
98.90 $\pm$ 0.20&
99.70 $\pm$ 0.08&
96.02 $\pm$ 0.18&
98.95 $\pm$ 0.09 \\
20W-1S&
\emph{Sum12} Duplicated&
\textbf{98.94 $\pm$ 0.19}&
\textbf{99.79 $\pm$ 0.05}&
96.45 $\pm$ 0.17&
99.07 $\pm$ 0.08 \\
\hline
20W-5S&
\emph{Sum12}&
98.37 $\pm$ 0.23&
99.69 $\pm$ 0.09&
95.11 $\pm$ 0.20&
98.94 $\pm$ 0.08 \\
20W-5S&
\emph{Sum12} Reduced&
\textbf{98.56 $\pm$ 0.19}&
99.71 $\pm$ 0.09&
\textbf{95.31 $\pm$ 0.19}&
98.86 $\pm$ 0.08 \\
20W-5S&
\emph{Sum12} Duplicated&
98.46 $\pm$ 0.21&
\textbf{99.74 $\pm$ 0.08}&
95.25 $\pm$ 0.19&
\textbf{99.07 $\pm$ 0.07} \\
\hline
\end{tabular}
\end{table*}

\begin{table*}[t]
\caption{\textbf{Comparison of \emph{Sum12} variants on CIFAR-FS. Classification accuracy (\%).}}
\label{tab:cifarfs_sum12_variants_duplicated_reduced}
\centering
\setlength{\tabcolsep}{3pt}
\begin{tabular}{|p{55pt}|p{90pt}|p{75pt}|p{75pt}|p{75pt}|p{75pt}|}
\hline
Trained at&
Encoder&
5W-1S&
5W-5S&
20W-1S&
20W-5S \\
\hline
5W-1S&
\emph{Sum12}&
62.82 $\pm$ 0.75&
77.46 $\pm$ 0.57&
35.42 $\pm$ 0.34&
53.18 $\pm$ 0.30 \\
5W-1S&
\emph{Sum12} Reduced&
61.59 $\pm$ 0.78&
76.32 $\pm$ 0.57&
33.75 $\pm$ 0.31&
51.39 $\pm$ 0.30 \\
5W-1S&
\emph{Sum12} Duplicated&
\textbf{63.22 $\pm$ 0.75}&
\textbf{77.72 $\pm$ 0.57}&
\textbf{35.76 $\pm$ 0.34}&
\textbf{53.64 $\pm$ 0.30} \\

\rowcolor{yellow}
5W-5S&
\emph{Sum12}&
62.25 $\pm$ 0.70&
80.20 $\pm$ 0.53&
35.20 $\pm$ 0.31&
57.47 $\pm$ 0.29 \\

5W-5S&
\emph{Sum12} Reduced&
63.25 $\pm$ 0.74&
81.50 $\pm$ 0.50&
35.81 $\pm$ 0.33&
58.91 $\pm$ 0.29 \\
5W-5S&
\emph{Sum12} Duplicated&
\textbf{65.13 $\pm$ 0.72}&
\textbf{82.38 $\pm$ 0.49}&
\textbf{37.99 $\pm$ 0.33}&
\textbf{61.01 $\pm$ 0.29} \\
\hline
20W-1S&
\emph{Sum12}&
\textbf{71.93 $\pm$ 0.77}&
\textbf{84.36 $\pm$ 0.47}&
45.81 $\pm$ 0.38&
63.41 $\pm$ 0.32 \\
20W-1S&
\emph{Sum12} Reduced&
70.07 $\pm$ 0.78&
82.28 $\pm$ 0.51&
43.47 $\pm$ 0.38&
61.18 $\pm$ 0.32 \\
20W-1S&
\emph{Sum12} Duplicated&
71.89 $\pm$ 0.72&
84.11 $\pm$ 0.48&
\textbf{46.16 $\pm$ 0.38}&
\textbf{64.21 $\pm$ 0.33} \\
\hline
20W-5S&
\emph{Sum12}&
\textbf{64.43 $\pm$ 0.74}&
\textbf{84.05 $\pm$ 0.45}&
\textbf{39.59 $\pm$ 0.38}&
64.73 $\pm$ 0.31 \\
20W-5S&
\emph{Sum12} Reduced&
61.99 $\pm$ 0.73&
83.10 $\pm$ 0.48&
36.53 $\pm$ 0.36&
62.95 $\pm$ 0.30 \\
20W-5S&
\emph{Sum12} Duplicated&
63.99 $\pm$ 0.74&
83.65 $\pm$ 0.46&
38.63 $\pm$ 0.36&
\textbf{64.75 $\pm$ 0.29} \\
\hline
\end{tabular}
\end{table*}

\subsubsection{Comparison with the Inception Module}\label{sec:Inception_ablation}

\begin{figure*}[!htbp]
  \centering
  \includegraphics[width=0.85\textwidth,height=0.66\textheight,keepaspectratio]{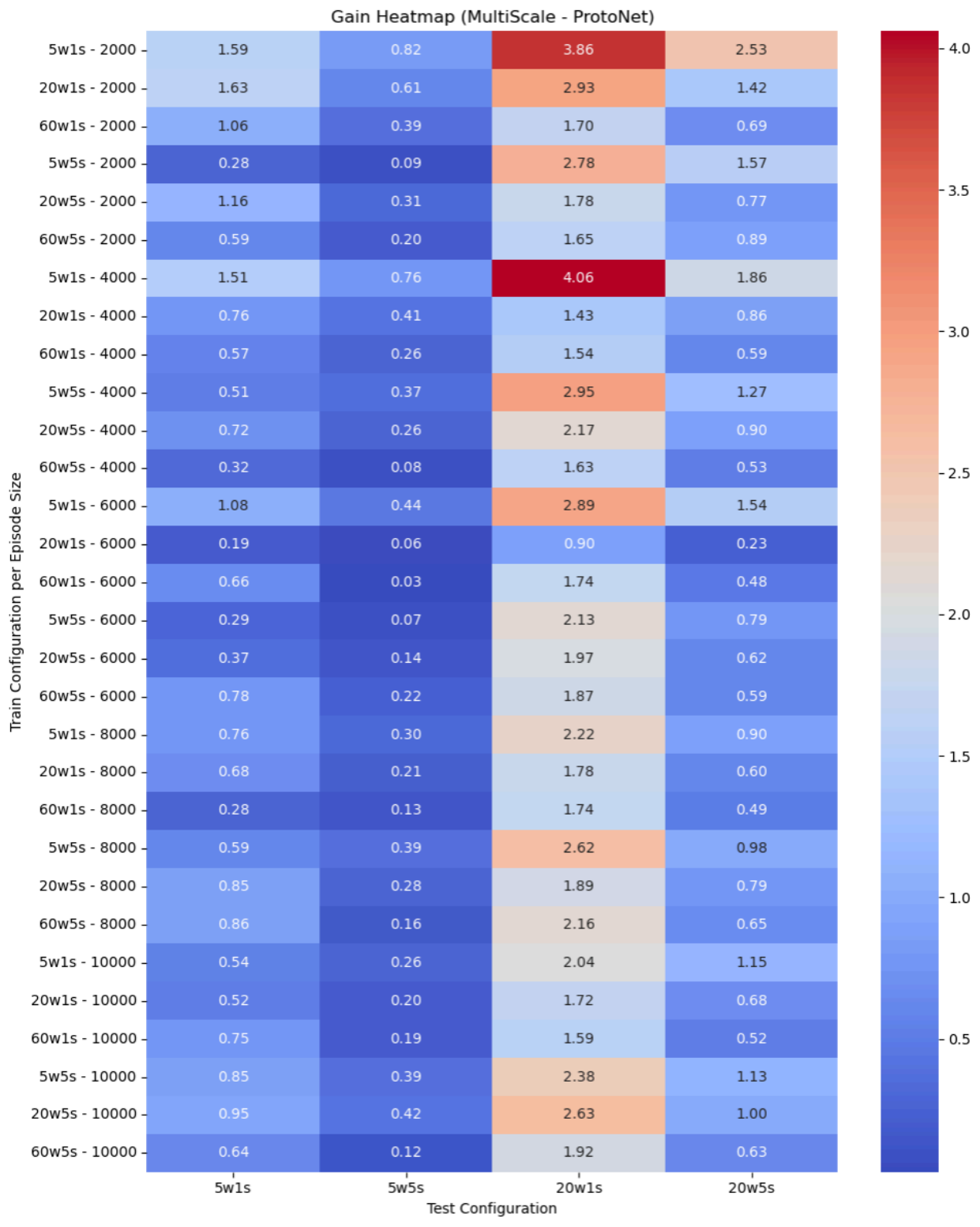}
    \caption{Gain heatmap comparing the proposed \textbf{Sum12 Multi-Scale encoder} against the standard Prototypical Network encoder. 
    Each cell represents the percentage gain in accuracy across different training configurations (rows) and test configurations (columns). 
    The color scale indicates relative improvement, with darker red indicating greater gains. The results highlight that Sum12 achieves 
    the largest improvements under high-way, low-shot regimes, especially in the 20-way 1-shot configuration.}
  \label{fig:gainheatmap}
\end{figure*}

\mbox{Table~\ref{tab:omniglot_full}} and \mbox{Figure~\ref{fig:gainheatmap}} show that, on Omniglot, both encoders reach saturation quickly, resulting in only small performance differences. By contrast, \mbox{Table~\ref{tab:fc100}} shows that the 20-way setups on FC100 remain far from saturation. Based on these observations, we selected CIFAR-FS as the most suitable dataset for evaluating architectural variants and comparing them with the Inception module. 

\hl{As mentioned in Section~\ref{sec:architecture}, one of the distinctions} between the Inception Module and the Sum12 Multi-Scale Branching encoder is the inclusion of pooling operations. The Inception Module dedicates an entire branch to max pooling, whose output is later concatenated with the outputs of the convolutional branches. By contrast, the Sum12 Encoder relies exclusively on the representational power of its convolutional blocks.

\begin{figure}[!t]
  \centering
  \includegraphics[width=0.72\linewidth]{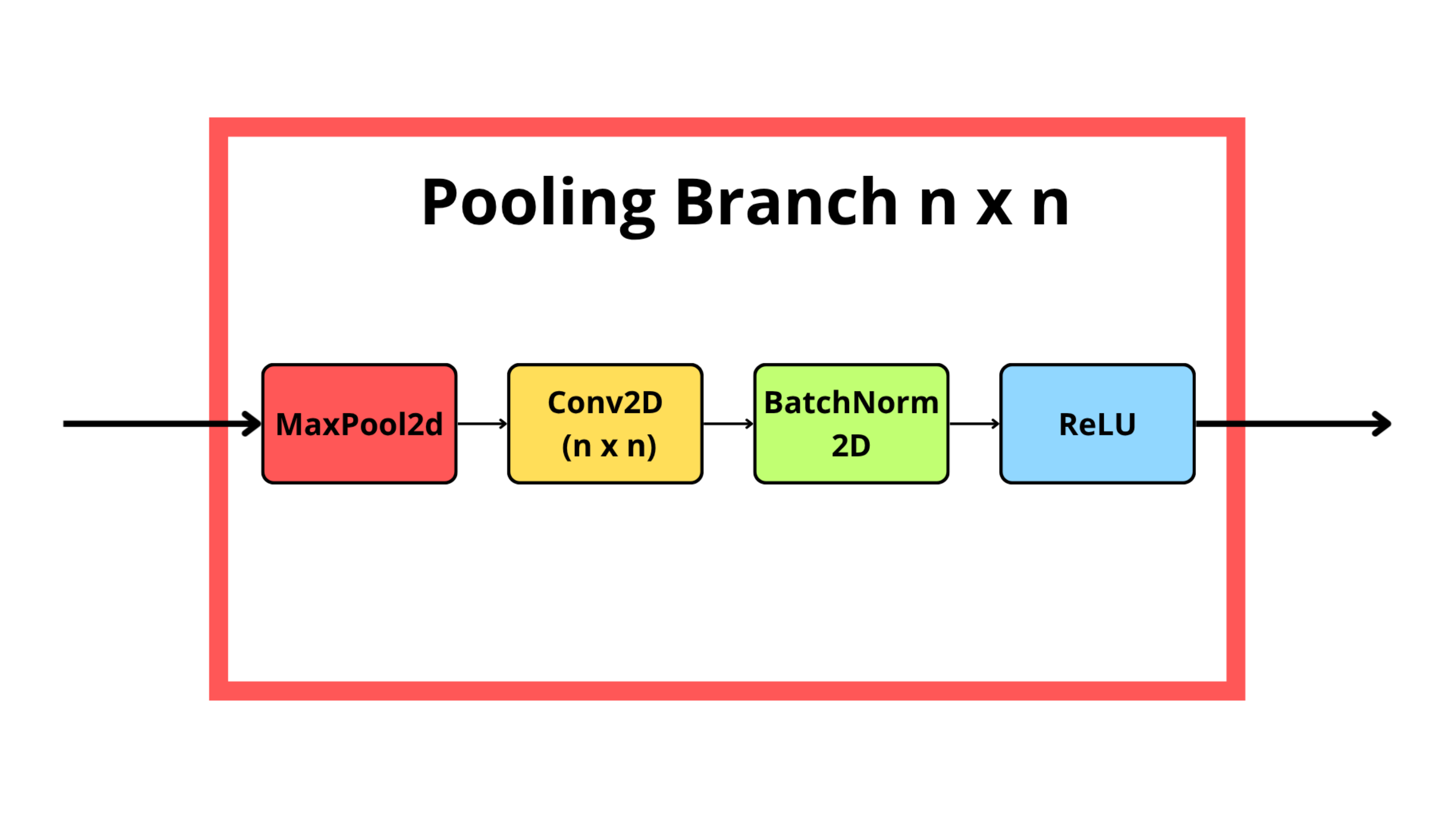}
  \caption{Pooling branch used in the Sum12 variant with pooling.}
  \label{fig:pooling_block}
\end{figure}

\begin{figure*}[!t]
  \centering
  \includegraphics[width=0.92\textwidth,height=0.64\textheight,keepaspectratio]{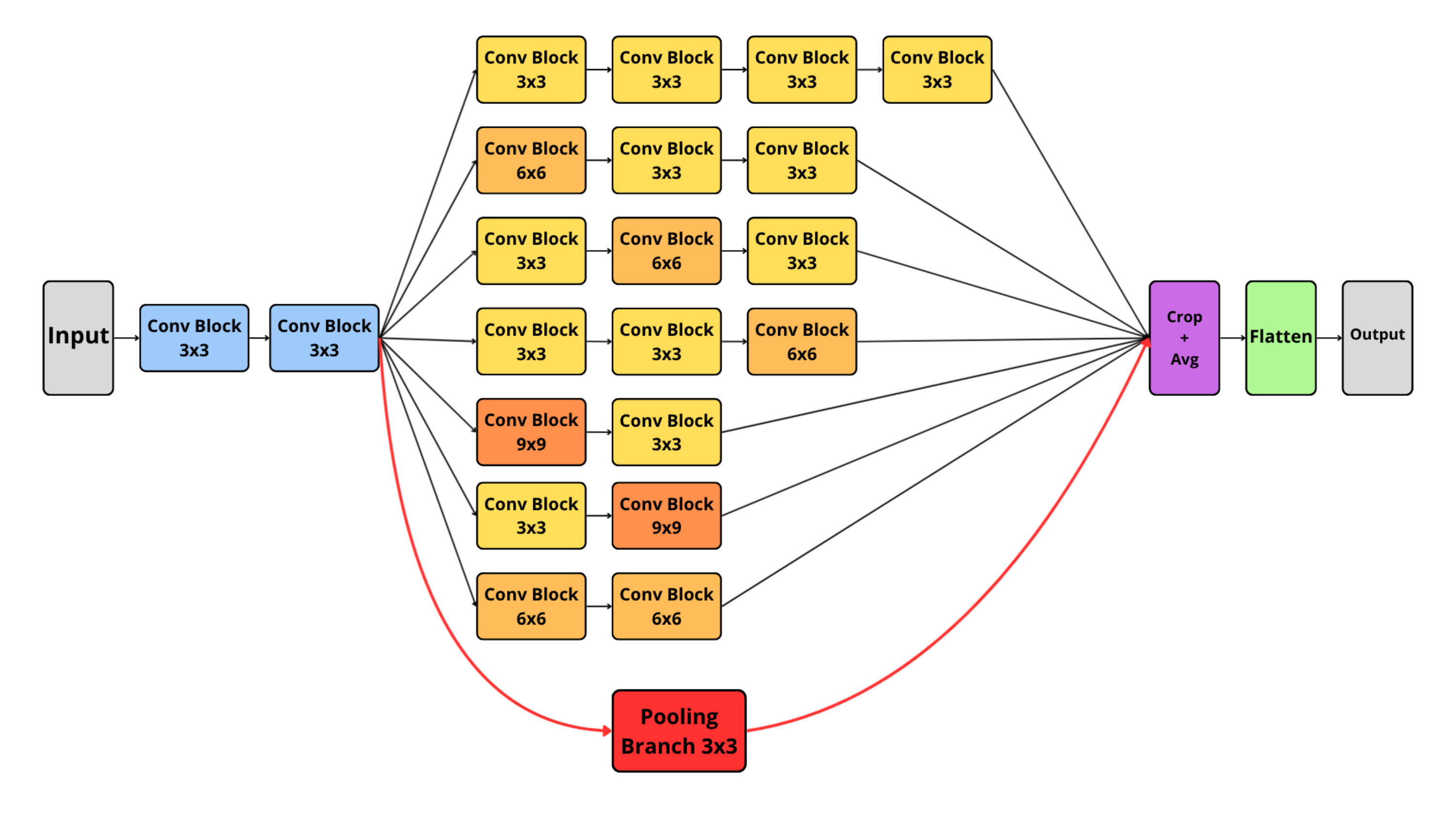}
  \caption{\emph{Sum12} encoder with an additional pooling branch.}
  \label{fig:sum12_w_Pool}
\end{figure*}

To rigorously assess the effectiveness of both architectures, we evaluated not only their standard forms but also two additional variants: an Inception Module without a pooling branch and a modified Sum12 Encoder that integrates pooling. \hl{This way the effect of an added pooling branch over the base encoder can be observed more clearly. To align} max pooling with the design philosophy of the Sum12 architecture, we defined a Pooling Block as a sequence comprising a max-pooling layer followed by an \(n{\times}n\) convolution, batch normalization, and ReLU activation, as illustrated in Figure~\ref{fig:pooling_block}. This Pooling Block was then incorporated as an additional parallel branch along the existing convolutional paths, as shown in Figure~\ref{fig:sum12_w_Pool}.

The results of these comparisons are listed in
\mbox{Table~\ref{tab:sum12vsInceptionModule}}.
As in the previous experiments, the models were trained on both 5-Way and 20-Way setups with 1-Shot and 5-Shot configurations, and each trained model was subsequently evaluated across all these scenarios. In all cases, using either the Inception Module or the Sum12 Encoder yielded a \hl{clear} improvement in accuracy over the baseline Prototypical Networks.

\hl{According to the reported results, the Sum12 Encoder with added pooling ranks first in classification accuracy in 7 out of 16 configurations, followed by the Inception Module, which ranks first in 4 out of 16 configurations, and Inception without pooling, which ranks first in 3 out of 16 configurations. The basic Sum12 Encoder ranks first in the remaining 2 configurations. When the models are trained and evaluated under the same }\(n\)\hl{-Way }\(k\)\hl{-Shot configuration, the Sum12 Multi-Scale Encoder with added pooling slightly outperforms the other variants, except in the 5-Way 1-Shot case when the opposite happens.

We do not claim that the Sum12 architecture is absolutely better than the Inception module. The standard Inception Module outperforms the basic Sum12 Multi-Scale Encoder in 14 of 16 cells, with two exceptions. The difference between Sum12 with pooling and the Inception module is narrow in almost all cases except for when the networks are trained on the 20-Way 5-Shot configuration and evaluated at the two 1-Shot scenarios, where the Inception module shows an advantage of 3.73\% and 3.70\% for 5-Way and 20-Way, respectively, over the Sum12 encoder with pooling branch.}

\subsection{MedMNIST 2D}\label{sec:medmnist2d_results}

In this section, we report the results achieved by the Sum12 Multi-Scale Branching encoder on the MedMNIST 2D benchmark and compare them with those of the original MedMNIST benchmark.

First, BreastMNIST is discussed separately because it \hl{exhibited} a behavioral pattern during training that was not observed in the other MedMNIST 2D datasets. As shown in \mbox{Table~\ref{tab:breastmnist}}, BreastMNIST exhibited high variance in the test results. Across the experiments conducted in this study, including the FSL experiments on
Omniglot, FC100, and CIFAR-FS, as well as the non-FSL experiments on MedMNIST 2D, this phenomenon was observed only for BreastMNIST. \hl{One possible explanation is the limited number of examples available in the dataset.}

\hl{Another factor that may contribute to this high variance is the class imbalance of BreastMNIST. As reported in \cite{medmnist}, BreastMNIST is derived from the BUSI dataset introduced in \cite{busiDataset} and consists of 780 breast ultrasound images. According to the corrected class counts reported in \cite{zhangTumorSegmentation}, the dataset contains 133 normal, 437 benign, and 210 malignant images. In BreastMNIST, the normal and benign categories are combined into a single non-malignant class, resulting in 570 non-malignant images and 210 malignant images. The resulting binary classification dataset is divided into 546 training, 78 validation, and 156 test samples \cite{medmnist}. 

The combination of the limited dataset size and the under-representation of the malignant class may increase sensitivity to initialization and sampling, thereby contributing to the observed variation across runs.}

\hl{For BreastMNIST, training was performed only at a resolution of }
\(28 \times 28\)\hl{, and the reported results were obtained from
50 independent runs. The mean AUC was }
\(0.8955 \pm 0.0209\)\hl{, with an observed range of }
\(0.8746\text{--}0.9164\)\hl{, whereas the mean ACC was }
\(0.8199 \pm 0.0882\)\hl{, with an observed range of }
\(0.7317\text{--}0.9081\)\hl{.}

We now discuss the performance of the Sum12 Branching Encoder on the remaining datasets of the MedMNIST 2D benchmark. Because encoder performance is evaluated here in non-FSL scenarios, and because the authors of \cite{medmnist} reported ResNet-18 and ResNet-50 performance separately for 28$\times$28 and 224$\times$224 resolutions, we follow the same convention. The full numerical values of the experiments are listed in \mbox{Table~\ref{tab:medmnist_all_panels}} of Appendix~\ref{app:medmnist}.

The results show that the Sum12 Multi-Scale encoder performs competitively relative to benchmark methods. In terms of classification accuracy (ACC), the encoder ranked first on \hl{three} datasets and second on three datasets when trained with 28$\times$28 images, and first on four datasets and second on three datasets when trained with 224$\times$224 images. The first- and second-place rankings for both AUC and ACC are summarized in \mbox{Table~\ref{tab:medmnist_rank_summary}}. With respect to ACC, the Sum12 Multi-Scale Encoder at 224$\times$224 performs better than any of the benchmark methods, ranking first on four datasets. Even at 28$\times$28, it compares favorably with Google AutoML Vision: both rank first on three datasets, \hl{and both rank second on three datasets.}

In terms of the area under the curve (AUC), Google AutoML Vision achieved the strongest overall performance, ranking first on four datasets, as summarized in \mbox{Table~\ref{tab:medmnist_rank_summary}}. The two Sum12 Multi-Scale variants closely follow this trend: the \(224{\times}224\) configuration ranks first on three datasets, while the \(28{\times}28\) configuration ranks first on two datasets. Both configurations outperform all the remaining benchmark methods, which rank first on at most one dataset or none. \mbox{Table~\ref{tab:medmnist_avg}} further reflects this competitive behavior in the average performance results, where the Sum12 Multi-Scale model trained at \(224{\times}224\) achieved an average AUC of \hl{91.9\%}, \hl{while Google AutoML Vision achieved the highest average AUC of} \(92.7\%\). These results indicate that, although Google AutoML Vision attained the highest peak AUC performance across datasets, the proposed Sum12 Multi-Scale Encoder remained consistently competitive and offered a favorable balance between top-ranked results and strong average performance.

\begin{table}
\caption{\textbf{Performance comparison on BreastMNIST in terms of AUC and accuracy (ACC). Results for the \emph{Sum12} Multi-Scale Encoder are reported as mean $\pm$ standard deviation, with the observed range over 50 \hl{independent runs} in brackets.}}
\label{tab:breastmnist}
\setlength{\tabcolsep}{3pt}
\begin{tabular}{|p{105pt}|p{55pt}|p{55pt}|}
\hline
Model&
AUC&
ACC \\
\hline
ResNet-18 (28)&
0.901&
0.863 \\
ResNet-18 (224)&
0.891&
0.833 \\
ResNet-50 (28)&
0.857&
0.812 \\
ResNet-50 (224)&
0.866&
0.842 \\
Auto-sklearn&
0.836&
0.803 \\
AutoKeras&
0.871&
0.831 \\
Google AutoML Vision&
0.919&
0.861 \\
\emph{Sum12} Multi-Scale Encoder&
0.8955 $\pm$ 0.0209 \par [0.8746--0.9164]&
0.8199 $\pm$ 0.0882 \par [0.7317--0.9081] \\
\hline
\end{tabular}
\end{table}

\begin{table}
\caption{\textbf{Ranking summary across MedMNIST v2 datasets. Counts reflect how many times each model ranked first or second across AUC and ACC. \hl{Ties are counted for each corresponding rank.} Full per-dataset results appear in \mbox{Table~\ref{tab:medmnist_all_panels}} in Appendix~\ref{app:medmnist}.}}
\label{tab:medmnist_rank_summary}
\setlength{\tabcolsep}{3pt}
\begin{tabular}{|p{105pt}|p{30pt}|p{30pt}|p{30pt}|p{30pt}|}
\hline
Model&
AUC 1st&
AUC 2nd&
ACC 1st&
ACC 2nd \\
\hline
ResNet-18 (28)&
0&
\hl{3}&
1&
\hl{1} \\
ResNet-18 (224)&
1&
\hl{\textbf{5}}&
1&
\hl{\textbf{3}} \\
ResNet-50 (28)&
1&
\hl{1}&
0&
\hl{\textbf{3}} \\
ResNet-50 (224)&
\hl{1}&
\hl{2}&
1&
1 \\
Auto-sklearn&
0&
0&
0&
0 \\
AutoKeras&
1&
\hl{1}&
2&
0 \\
Google AutoML Vision&
\textbf{4}&
\hl{2}&
\hl{3}&
\hl{\textbf{3}} \\
\emph{Sum12} Multi-Scale (28)&
2&
\hl{\textbf{5}}&
\hl{3}&
\hl{\textbf{3}} \\
\emph{Sum12} Multi-Scale (224)&
3&
\hl{2}&
\hl{\textbf{4}}&
\hl{\textbf{3}} \\
\hline
\end{tabular}
\end{table}

\begin{table}
\caption{\textbf{Average performance across all MedMNIST v2 datasets in terms of AUC and ACC. Full per-dataset results are reported in \mbox{Table~\ref{tab:medmnist_all_panels}} in Appendix~\ref{app:medmnist}.}}
\label{tab:medmnist_avg}
\setlength{\tabcolsep}{3pt}
\begin{tabular}{|p{110pt}|p{55pt}|p{55pt}|}
\hline
Method&
Average AUC&
Average ACC \\
\hline
ResNet-18 (28) \cite{medmnist}&
0.922&
0.819 \\
ResNet-18 (224) \cite{medmnist}&
\textit{0.925}&
0.822 \\
ResNet-50 (28) \cite{medmnist}&
0.920&
0.816 \\
ResNet-50 (224) \cite{medmnist}&
0.923&
0.821 \\
Auto-sklearn \cite{medmnist}&
0.878&
0.722 \\
AutoKeras \cite{medmnist}&
0.917&
0.813 \\
Google AutoML Vision \cite{medmnist}&
\textbf{0.927}&
0.809 \\
\emph{Sum12} Multi-Scale (28)&
0.923&
\textit{0.832} \\
\emph{Sum12} Multi-Scale (224)&
0.919&
\textbf{0.834} \\
\hline
\end{tabular}
\end{table}

\subsection{STM32 Deployment}\label{sec:section_stm32_deployment}
\hl{Encoders trained under a standard training regime on MedMNIST 2D also retain their lightweight nature, with a model size of approximately 7{,}660\,kB, only slightly larger than the 7{,}426\,kB models trained on Omniglot.} The small model size of these networks, together with the fact that they were built almost exclusively from convolutional and simple matrix operations, prompted us to test their deployment on microcontrollers. 

The model was converted from \emph{.pth} format to \emph{.onnx} format and imported into the \hl{STM32 AI Studio and the STM32CubeIDE} \hl{with the aim of deploying} it on an STM32 microcontroller board. The IDE recognized the trained model, as shown in Figure~\ref{fig:stm32graph_ide} in Appendix~\ref{app:stm32deployment}, which presents the graph produced by the STM32CubeIDE during successful model recognition. \hl{The model was still too large} for deployment on boards relying only on the internal memory of the STM32H753ZI MCU, such as the STM32 NUCLEO-H753ZI, which provides 2 MB of embedded flash memory and 1 MB of RAM.

\hl{Since the dilated Sum12 variant is substantially lighter and, when trained under the 5-Way 5-Shot configuration, outperforms the standard Sum12 encoder in three out of four evaluation settings reported in Table~\ref{tab:cifarfs_dilatation}, we selected this variant for deployment on the STM32 NUCLEO-H753ZI microcontroller board
. Although the \emph{.onnx} model has a size of 2{,}904~kB, the STM32 AI Studio evaluated a flash-memory usage of 2{,}892~kB for weights and a RAM usage of 270~kB for activations.

STM32 AI Studio identified the convolutional weight arrays as the principal memory consumers. Therefore, we initially applied partial INT8 quantization to these components while retaining FP32 precision elsewhere. Other quantizations such as 16-bit Floating Point (FP16) are not supported by the STM Software.

This quantization reduced the on-disk model size to 1{,}642~kB, which the STM32 AI Studio estimated as requiring 2{,}028~kB of flash memory and 278~kB of RAM, as shown in Figure~\ref{fig:stm32_partial_quantization} of Appendix~\ref{app:stm32deployment}. Although this variant with partial quantization was validated successfully on the host computer, it was not possible to deploy it on the target board. The IDE log reported the error \emph{region `FLASH' overflowed by 143{,}612 bytes}, as the hardware usage did not include the additional code and data required by the firmware.

Consequently, we applied full INT8 quantization, which reduced the on-disk size of the \emph{.onnx} model to 799\,kB. The resulting model was successfully deployed on the NUCLEO-H753ZI board and validated using 100 CIFAR-FS images converted to \emph{.npy} format for compatibility with STM32 AI Studio. The hardware usage of this deployment on the target board is shown in Figure~\ref{fig:stm32_on_target} of Appendix~\ref{app:stm32deployment}. 

The original FP32 model, the partially quantized model, and the fully INT8-quantized \emph{.onnx} model were also evaluated in the native Python environment. Their accuracies fell within the range reported in Table~\ref{tab:cifarfs_dilatation} when both training and evaluation used the 5-Way 5-Shot regime. Both the mean and minimum cosine similarities of the partially and fully INT8-quantized models were approximately $99.99\%$ relative to the original FP32 model, with largest observed errors of $1.7\%$ and $3.7\%$, respectively, as shown in Table~\ref{tab:sum12_quantization_embedding}. The validation of the C implementation generated by the STM32 AI Studio on the target board produced a substantially lower cosine similarity of approximately $35.2\%$ and a mean absolute error of approximately $28.2\%$. Because this validation was performed automatically by the IDE, it did not report classification accuracy.

These results show that full INT8 quantization enables the dilated Sum12 model to fit within the memory budget of the target microcontroller. However, the low cosine similarity and high error observed on the target indicate that the generated embedded implementation has not yet been functionally validated for few-shot classification.

}

\begin{table}[t]
\caption{\hl{\textbf{Embedding preservation comparison for the partially quantized and fully INT8-quantized \emph{Sum12} Multi-Scale Encoder with Dilated Convolutions}}}
\label{tab:sum12_quantization_embedding}
\centering
\begingroup
\rowcolors{1}{yellow}{yellow}

\setlength{\tabcolsep}{3pt}
\begin{tabular}{|p{105pt}|p{55pt}|p{55pt}|}
\hline
Metric&
Partial INT8&
Full INT8 \\
\hline
Mean cosine similarity&
0.9999795&
0.9999188 \\
Minimum cosine similarity&
0.9999675&
0.9998547 \\
Mean absolute error&
0.001894&
0.003835 \\
Largest observed error&
0.01704&
0.03707 \\
\hline
\end{tabular}
\endgroup

\end{table}

\section{Conclusions and Future Work}\label{sec:conclusion}

Overall, this paper presents a novel architecture that performs \hl{well relative to other backbone architectures} in both standard supervised learning scenarios and Few-Shot Learning (FSL) settings, where it serves as an encoder for prototype computation.

The proposed architecture was evaluated on 16 datasets across both few-shot and conventional supervised learning settings. \hl{In the controlled FSL comparisons, it generally performed better than the vanilla ProtoNet encoder, making it well-suited for prototype learning.} Furthermore, it achieved a performance comparable \hl{to that of existing} methods on Omniglot and can be extended to \hl{compete favorably with} the Inception module in selected FSL configurations. We also demonstrated that, in 5-way 1-shot and 5-way 5-shot scenarios, the Sum12 Multi-Scale encoder increases the performance of Prototypical Networks on FC100 into the lower bound of the accuracy range reported by established FSL methods. Based on the results obtained on Omniglot, CIFAR-FS, and FC100, future work may investigate how this encoder, or derived architectures, can be adapted to unsupervised learning scenarios or combined with more advanced methods such as CLIP~\cite{clip_method}.

Within the MedMNIST 2D collection of 12 medical datasets, the Sum12 Multi-Scale encoder, evaluated at $224 \times 224$ resolution, achieved higher accuracy across more datasets than the benchmark models. This behavior suggests that the proposed encoder maintains stable classification performance across heterogeneous medical imaging tasks with varying modalities and anatomical targets. Among the phenomena observed in the MedMNIST 2D investigation, the high variance and apparent instability of the BreastMNIST dataset, which is composed of breast ultrasound images, highlight the limitations of the current training regimes on small and imbalanced medical datasets, suggesting a promising direction for future work aimed at improving the applicability of few-shot learning methods to breast ultrasound and mammography imaging. A further research direction emerging from the MedMNIST 2D results is the exploration of 3D-capable Multi-Scale branching encoders, particularly for volumetric medical imaging data.

\hl{Although the Sum12 Multi-Scale Branching Encoder performs competitively against methods based on basic ProtoNet, Matching Networks, Relation Networks, and Meta-Learner LSTM, it remains behind deeper architectures such as ResNet-12 and WRN-28-10, particularly when those backbones are combined with state-of-the-art methods.}

\hl{Furthermore, the network is relatively lightweight compared with many standard convolutional backbones, and its size was reduced further by replacing standard convolutions with dilated convolutions and applying partial or full quantization. Among these variants, the fully INT8-quantized dilated model fit within the memory budget of the STM32 NUCLEO-H753ZI board. However, the low cosine similarity and high mean absolute error observed for the generated on-target implementation indicate that it has not yet been functionally validated for few-shot classification. Future work should determine why the embeddings produced by the STM32-generated C implementation differ from those produced by the ONNX model in Python and should implement an embedded few-shot evaluation pipeline tailored to the Sum12 encoder.}

\hl{Another direction for future work is the evaluation of the Sum12 encoder through repeated independent training runs and formal statistical significance analysis. Formal statistical significance testing was not performed in the present study and therefore remains explicitly outside the scope of the current experimental evaluation. In future few-shot learning experiments, the competing encoders can be evaluated on identical sets of sampled episodes to enable paired comparisons and more reliable uncertainty estimation, as recommended by \cite{lafargue2024}. Furthermore, non-parametric procedures can also be considered~\cite{demsar2006}. Likewise, t-distributed Stochastic Neighbor Embedding (t-SNE)~\cite{vandermaaten2008} and Uniform Manifold Approximation and Projection (UMAP)~\cite{mcinnes2018} were not performed in the present study and remain directions for future work. These methods can be used to examine the feature representations learned by the baseline and proposed encoders and may provide further qualitative insight into intra-class compactness and inter-class separation for previously unseen categories.}

\clearpage        

\appendices

\section{Extended Omniglot Results}\label{app:omniglot_extended_results}

\begin{table*}[!b]
\caption{\textbf{Comparison of ProtoNet and \emph{Sum12} Multi-Scale encoders across training episodes on the Omniglot dataset. Each row corresponds to a specific training configuration, and evaluation results are reported for 5-Way and 20-Way tasks with both 1-Shot and 5-Shot support sets.}}
\label{tab:omniglot_full}
\centering
\small
\setlength{\tabcolsep}{3pt}
\begin{tabular}{|p{28pt}|p{42pt}|p{38pt}|p{38pt}|p{38pt}|p{38pt}|p{38pt}|p{38pt}|p{38pt}|p{38pt}|}
\hline
N.eps&
Trained at&
\multicolumn{2}{p{76pt}|}{5W-1S}&
\multicolumn{2}{p{76pt}|}{5W-5S}&
\multicolumn{2}{p{76pt}|}{20W-1S}&
\multicolumn{2}{p{76pt}|}{20W-5S} \\
\hline
&
&
ProtoNet&
\emph{Sum12}&
ProtoNet&
\emph{Sum12}&
ProtoNet&
\emph{Sum12}&
ProtoNet&
\emph{Sum12} \\
\hline
2k & 5W-1S  & 93.14 & \textbf{94.73} & 97.69 & \textbf{98.51} & 82.06 & \textbf{85.92} & 93.11 & \textbf{95.64} \\
2k & 20W-1S & 96.68 & \textbf{98.31} & 98.96 & \textbf{99.57} & 91.54 & \textbf{94.47} & 97.02 & \textbf{98.44} \\
2k & 60W-1S & 97.48 & \textbf{98.54} & 99.31 & \textbf{99.70} & 93.65 & \textbf{95.35} & 98.00 & \textbf{98.69} \\
2k & 5W-5S  & 92.73 & \textbf{93.01} & 98.17 & \textbf{98.26} & 81.01 & \textbf{83.79} & 93.74 & \textbf{95.31} \\
2k & 20W-5S & 96.19 & \textbf{97.35} & 99.06 & \textbf{99.37} & 90.14 & \textbf{91.92} & 97.25 & \textbf{98.02} \\
2k & 60W-5S & 96.96 & \textbf{97.55} & 99.37 & \textbf{99.57} & 92.02 & \textbf{93.67} & 97.73 & \textbf{98.62} \\
\hline
4k & 5W-1S  & 95.32 & \textbf{96.83} & 98.45 & \textbf{99.21} & 86.64 & \textbf{90.70} & 95.51 & \textbf{97.37} \\
4k & 20W-1S & 97.73 & \textbf{98.49} & 99.29 & \textbf{99.70} & 93.44 & \textbf{94.87} & 97.80 & \textbf{98.66} \\
4k & 60W-1S & 98.29 & \textbf{98.86} & 99.52 & \textbf{99.78} & 94.61 & \textbf{96.15} & 98.44 & \textbf{99.03} \\
4k & 5W-5S  & 94.75 & \textbf{95.26} & 98.72 & \textbf{99.09} & 84.71 & \textbf{87.66} & 95.76 & \textbf{97.03} \\
4k & 20W-5S & 97.15 & \textbf{97.87} & 99.32 & \textbf{99.58} & 91.58 & \textbf{93.75} & 97.82 & \textbf{98.72} \\
4k & 60W-5S & 97.86 & \textbf{98.18} & 99.64 & \textbf{99.72} & 93.08 & \textbf{94.71} & 98.36 & \textbf{98.89} \\
\hline
6k & 5W-1S  & 96.00 & \textbf{97.08} & 98.80 & \textbf{99.24} & 88.84 & \textbf{91.73} & 96.15 & \textbf{97.69} \\
6k & 20W-1S & 98.05 & \textbf{98.24} & 99.46 & \textbf{99.52} & 93.92 & \textbf{94.82} & 98.13 & \textbf{98.36} \\
6k & 60W-1S & 98.36 & \textbf{99.02} & 99.69 & \textbf{99.72} & 94.96 & \textbf{96.70} & 98.70 & \textbf{99.18} \\
6k & 5W-5S  & 95.39 & \textbf{95.68} & 98.88 & \textbf{98.95} & 87.24 & \textbf{89.37} & 96.50 & \textbf{97.29} \\
6k & 20W-5S & 97.83 & \textbf{98.20} & 99.55 & \textbf{99.69} & 92.94 & \textbf{94.91} & 98.41 & \textbf{99.03} \\
6k & 60W-5S & 97.64 & \textbf{98.42} & 99.48 & \textbf{99.70} & 93.43 & \textbf{95.30} & 98.47 & \textbf{99.06} \\
\hline
8k & 5W-1S  & 96.58 & \textbf{97.34} & 98.93 & \textbf{99.23} & 90.37 & \textbf{92.59} & 96.75 & \textbf{97.65} \\
8k & 20W-1S & 98.23 & \textbf{98.91} & 99.57 & \textbf{99.78} & 94.51 & \textbf{96.29} & 98.43 & \textbf{99.03} \\
8k & 60W-1S & 98.84 & \textbf{99.12} & 99.66 & \textbf{99.79} & 95.21 & \textbf{96.95} & 98.73 & \textbf{99.22} \\
8k & 5W-5S  & 96.29 & \textbf{96.88} & 98.93 & \textbf{99.32} & 89.26 & \textbf{91.88} & 97.01 & \textbf{97.99} \\
8k & 20W-5S & 97.40 & \textbf{98.25} & 99.43 & \textbf{99.71} & 93.24 & \textbf{95.13} & 98.19 & \textbf{98.98} \\
8k & 60W-5S & 97.78 & \textbf{98.64} & 99.56 & \textbf{99.72} & 93.53 & \textbf{95.69} & 98.49 & \textbf{99.14} \\
\hline
10k & 5W-1S  & 96.99 & \textbf{97.53} & 99.05 & \textbf{99.31} & 90.97 & \textbf{93.01} & 96.79 & \textbf{97.94} \\
10k & 20W-1S & 98.31 & \textbf{98.83} & 99.47 & \textbf{99.67} & 94.74 & \textbf{96.46} & 98.39 & \textbf{99.07} \\
10k & 60W-1S & 98.44 & \textbf{99.19} & 99.61 & \textbf{99.80} & 95.23 & \textbf{96.82} & 98.59 & \textbf{99.11} \\
10k & 5W-5S  & 96.29 & \textbf{97.14} & 99.14 & \textbf{99.53} & 89.75 & \textbf{92.13} & 97.01 & \textbf{98.14} \\
10k & 20W-5S & 97.41 & \textbf{98.36} & 99.31 & \textbf{99.73} & 92.93 & \textbf{95.56} & 98.04 & \textbf{99.04} \\
10k & 60W-5S & 97.93 & \textbf{98.57} & 99.59 & \textbf{99.71} & 93.76 & \textbf{95.68} & 98.57 & \textbf{99.20} \\
\hline
\end{tabular}
\end{table*}

\clearpage          

\section{Extended CIFAR-FS Results}\label{app:cifarfs_extended_results}

\begin{table*}[!b]
\caption{\textbf{Comparison of ProtoNet and \emph{Sum12} Multi-Scale encoders across training episodes on the CIFAR-FS dataset. Each row corresponds to a specific training configuration, and evaluation results are reported for 5-Way and 20-Way tasks with both 1-Shot and 5-Shot support sets.\hl{These results were obtained from an independently executed training-trajectory experiment and therefore do not correspond to the exact training runs reported in Table 4.}}}
\label{tab:cifarfs_extended}
\centering
\small
\setlength{\tabcolsep}{3pt}
\begin{tabular}{|p{28pt}|p{42pt}|p{38pt}|p{38pt}|p{38pt}|p{38pt}|p{38pt}|p{38pt}|p{38pt}|p{38pt}|}
\hline
N.eps&
Trained at&
\multicolumn{2}{p{76pt}|}{5W-1S}&
\multicolumn{2}{p{76pt}|}{5W-5S}&
\multicolumn{2}{p{76pt}|}{20W-1S}&
\multicolumn{2}{p{76pt}|}{20W-5S} \\
\hline
&
&
ProtoNet&
\emph{Sum12}&
ProtoNet&
\emph{Sum12}&
ProtoNet&
\emph{Sum12}&
ProtoNet&
\emph{Sum12} \\
\hline
2k & 5W-1S  & 44.10 & \textbf{45.71} & 58.55 & \textbf{61.01} & 19.93 & \textbf{21.42} & 32.40 & \textbf{34.56} \\
2k & 5W-5S  & 48.94 & \textbf{51.53} & 67.86 & \textbf{69.25} & 23.55 & \textbf{25.51} & 41.17 & \textbf{44.08} \\
2k & 20W-1S & 53.50 & \textbf{55.93} & 69.13 & \textbf{71.99} & 26.82 & \textbf{29.71} & 42.26 & \textbf{45.96} \\
2k & 20W-5S & 56.50 & \textbf{60.89} & 76.23 & \textbf{79.88} & 30.13 & \textbf{34.82} & 51.43 & \textbf{57.03} \\
\hline
4k & 5W-1S  & 47.20 & \textbf{52.64} & 62.85 & \textbf{69.03} & 22.12 & \textbf{26.44} & 35.81 & \textbf{42.67} \\
4k & 5W-5S  & 52.45 & \textbf{56.10} & 71.14 & \textbf{74.06} & 26.33 & \textbf{29.70} & 45.02 & \textbf{49.47} \\
4k & 20W-1S & 58.55 & \textbf{63.66} & 73.26 & \textbf{77.79} & 31.12 & \textbf{36.65} & 48.18 & \textbf{53.94} \\
4k & 20W-5S & 59.31 & \textbf{65.22} & 79.57 & \textbf{83.00} & 32.34 & \textbf{38.79} & 55.85 & \textbf{63.01} \\
\hline
6k & 5W-1S  & 52.58 & \textbf{56.59} & 68.28 & \textbf{71.93} & 26.33 & \textbf{29.44} & 41.87 & \textbf{45.61} \\
6k & 5W-5S  & 55.60 & \textbf{59.84} & 74.73 & \textbf{77.71} & 29.34 & \textbf{32.93} & 49.98 & \textbf{54.52} \\
6k & 20W-1S & 61.72 & \textbf{68.60} & 75.78 & \textbf{81.10} & 33.66 & \textbf{41.62} & 51.75 & \textbf{59.53} \\
6k & 20W-5S & 59.30 & \textbf{64.75} & 79.66 & \textbf{83.64} & 33.06 & \textbf{38.39} & 57.05 & \textbf{63.33} \\
\hline
8k & 5W-1S  & 54.25 & \textbf{60.35} & 69.31 & \textbf{75.44} & 26.62 & \textbf{33.07} & 43.08 & \textbf{50.21} \\
8k & 5W-5S  & 56.49 & \textbf{63.32} & 76.16 & \textbf{80.96} & 30.24 & \textbf{36.05} & 51.51 & \textbf{58.60} \\
8k & 20W-1S & 62.15 & \textbf{71.46} & 77.75 & \textbf{83.49} & 34.89 & \textbf{44.39} & 53.00 & \textbf{62.17} \\
8k & 20W-5S & 59.12 & \textbf{64.24} & 79.19 & \textbf{84.29} & 32.64 & \textbf{38.65} & 57.20 & \textbf{64.59} \\
\hline
10k & 5W-1S  & 56.91 & \textbf{62.84} & 72.10 & \textbf{77.52} & 28.87 & \textbf{34.86} & 45.87 & \textbf{52.65} \\
10k & 5W-5S  & 57.03 & \textbf{64.07} & 77.04 & \textbf{81.72} & 30.58 & \textbf{37.11} & 52.58 & \textbf{59.90} \\
10k & 20W-1S & 62.48 & \textbf{72.28} & 78.09 & \textbf{83.80} & 35.48 & \textbf{46.10} & 53.69 & \textbf{63.71} \\
10k & 20W-5S & 59.21 & \textbf{65.64} & 80.23 & \textbf{84.34} & 33.16 & \textbf{39.28} & 57.75 & \textbf{65.15} \\
\hline
\end{tabular}
\end{table*}

\clearpage          

\section{Sum12 Ablations}\label{app:sum12_ablations}

\begin{table*}[!b]
\caption{\textbf{Crop strategy exploration on Omniglot with Euclidean and cosine distances. Results report classification accuracy (\%).}}
\label{tab:omniglot_crop_strategy}
\centering
\small
\setlength{\tabcolsep}{2pt}
\begin{tabular}{|p{38pt}|p{88pt}|p{38pt}|p{38pt}|p{38pt}|p{38pt}|p{38pt}|p{38pt}|p{38pt}|p{38pt}|}
\hline
Trained at&
Crop Type&
\multicolumn{4}{p{152pt}|}{Euclidean}&
\multicolumn{4}{p{152pt}|}{Cosine} \\
\hline
&
&
5W-1S&
5W-5S&
20W-1S&
20W-5S&
5W-1S&
5W-5S&
20W-1S&
20W-5S \\
\hline
5W-1S& crop-collapse      & 97.60 & 99.16 & 92.52 & 97.38 & \textbf{86.95} & \textbf{92.04} & \textbf{69.05} & \textbf{78.50} \\
5W-1S& zero\_pad\_max     & 96.88 & 98.80 & 90.76 & 96.29 & 85.08 & 90.70 & 64.95 & 74.90 \\
5W-1S& same\_w\_averaging & \textbf{97.64} & \textbf{99.18} & \textbf{92.79} & \textbf{97.39} & 85.05 & 90.79 & 65.04 & 75.03 \\
5W-1S& adaptive           & 97.44 & 99.12 & 92.14 & 97.11 & 84.74 & 90.46 & 64.25 & 74.10 \\
\hline
5W-5S& crop-collapse      & \textbf{97.25} & \textbf{99.09} & \textbf{91.76} & \textbf{97.08} & \textbf{89.60} & \textbf{94.34} & \textbf{75.29} & \textbf{84.00} \\
5W-5S& zero\_pad\_max     & 97.06 & 99.03 & 90.82 & 96.68 & 88.22 & 93.28 & 71.97 & 81.55 \\
5W-5S& same\_w\_averaging & 97.12 & 99.00 & 91.47 & 96.91 & 88.16 & 93.22 & 71.94 & 81.25 \\
5W-5S& adaptive           & 97.03 & 98.95 & 91.49 & 96.80 & 88.27 & 93.42 & 72.53 & 81.96 \\
\hline
20W-1S& crop-collapse      & 98.89 & \textbf{99.69} & 96.21 & \textbf{98.86} & 81.85 & 89.03 & 63.93 & 71.87 \\
20W-1S& zero\_pad\_max     & 98.86 & 99.67 & 95.98 & 98.78 & \textbf{85.93} & \textbf{91.23} & \textbf{67.25} & \textbf{75.37} \\
20W-1S& same\_w\_averaging & \textbf{98.95} & 99.68 & 96.29 & 98.83 & 84.73 & 90.64 & 65.97 & 74.18 \\
20W-1S& adaptive           & 98.94 & 99.68 & \textbf{96.30} & 98.85 & 85.28 & 90.98 & 66.65 & 74.69 \\
\hline
20W-5S& crop-collapse      & 98.67 & 99.66 & 95.55 & 98.75 & 68.56 & 78.77 & 51.57 & 60.74 \\
20W-5S& zero\_pad\_max     & 98.60 & 99.62 & 95.18 & 98.55 & 76.05 & 85.04 & 58.30 & 67.55 \\
20W-5S& same\_w\_averaging & 98.75 & 99.67 & 95.65 & 98.76 & 68.11 & 78.25 & 51.87 & 61.02 \\
20W-5S& adaptive           & \textbf{98.79} & \textbf{99.69} & \textbf{95.76} & \textbf{98.82} & \textbf{76.87} & \textbf{85.43} & \textbf{58.55} & \textbf{67.62} \\
\hline
\end{tabular}
\end{table*}

\begin{table*}[!b]
\caption{\textbf{Performance comparison between ProtoNet, Inception, and the \emph{Sum12} Multi-Scale Encoder across different training and evaluation settings.}}
\label{tab:sum12vsInceptionModule}
\centering
\small
\setlength{\tabcolsep}{2pt}
\begin{tabular}{|p{48pt}|p{130pt}|p{72pt}|p{72pt}|p{72pt}|p{72pt}|}
\hline
Trained at&
Encoder&
\multicolumn{2}{p{144pt}|}{5-Way}&
\multicolumn{2}{p{144pt}|}{20-Way} \\
\hline
&
&
1-Shot&
5-Shot&
1-Shot&
5-Shot \\
\hline
5W-1S&
ProtoNet&
56.69 $\pm$ 0.73&
72.02 $\pm$ 0.60&
29.06 $\pm$ 0.29&
45.47 $\pm$ 0.30 \\
5W-1S&
\emph{Sum12} Multi-Scale&
62.82 $\pm$ 0.75&
77.46 $\pm$ 0.57&
\textbf{35.42 $\pm$ 0.34}&
\textbf{53.18 $\pm$ 0.30} \\
5W-1S&
Inception without Pooling&
61.97 $\pm$ 0.79&
75.77 $\pm$ 0.60&
33.32 $\pm$ 0.31&
49.81 $\pm$ 0.30 \\
5W-1S&
Inception Module&
\textbf{63.97 $\pm$ 0.80}&
77.57 $\pm$ 0.59&
35.14 $\pm$ 0.33&
51.43 $\pm$ 0.31 \\
5W-1S&
\emph{Sum12} Multi-Scale with Pooling&
63.33 $\pm$ 0.77&
\textbf{77.87 $\pm$ 0.55}&
35.23 $\pm$ 0.32&
53.07 $\pm$ 0.30 \\
\hline

\rowcolor{yellow}
5W-5S&
ProtoNet&
56.47 $\pm$ 0.76&
75.03 $\pm$ 0.58&
30.02 $\pm$ 0.32&
49.61 $\pm$ 0.29 \\

\rowcolor{yellow}
5W-5S&
\emph{Sum12} Multi-Scale&
62.25 $\pm$ 0.70&
80.20 $\pm$ 0.53&
35.20 $\pm$ 0.31&
57.47 $\pm$ 0.29 \\

\rowcolor{yellow}
5W-5S&
Inception without Pooling&
64.79 $\pm$ 0.75&           %
80.51 $\pm$ 0.54&           %
36.80 $\pm$ 0.34&           %
56.64 $\pm$ 0.31 \\         %

\rowcolor{yellow}
5W-5S&
Inception Module&
\textbf{65.44 $\pm$ 0.74}&           %
81.49 $\pm$ 0.53&           %
\textbf{37.93 $\pm$ 0.36}&           %
57.55 $\pm$ 0.30 \\         %

\rowcolor{yellow}
5W-5S&
\emph{Sum12} Multi-Scale with Pooling&
62.37 $\pm$ 0.74&           %
\textbf{81.65 $\pm$ 0.50}&           %
36.29 $\pm$ 0.33&           %
\textbf{59.41 $\pm$ 0.29} \\         %
\hline
20W-1S&
ProtoNet&
64.16 $\pm$ 0.79&
78.39 $\pm$ 0.57&
36.21 $\pm$ 0.33&
54.37 $\pm$ 0.32 \\
20W-1S&
\emph{Sum12} Multi-Scale&
71.93 $\pm$ 0.77&
84.36 $\pm$ 0.47&
45.81 $\pm$ 0.38&
63.41 $\pm$ 0.32 \\
20W-1S&
Inception without Pooling&
\textbf{73.61 $\pm$ 0.79}&
84.27 $\pm$ 0.50&
46.63 $\pm$ 0.41&
63.98 $\pm$ 0.33 \\
20W-1S&
Inception Module&
72.84 $\pm$ 0.78&
84.65 $\pm$ 0.51&
47.39 $\pm$ 0.39&
64.41 $\pm$ 0.32 \\
20W-1S&
\emph{Sum12} Multi-Scale with Pooling&
72.05 $\pm$ 0.77&
\textbf{84.67 $\pm$ 0.50}&
\textbf{47.52 $\pm$ 0.39}&
\textbf{65.46 $\pm$ 0.31} \\
\hline
20W-5S&
ProtoNet&
58.84 $\pm$ 0.76&
80.42 $\pm$ 0.52&
32.63 $\pm$ 0.34&
57.82 $\pm$ 0.31 \\
20W-5S&
\emph{Sum12} Multi-Scale&
64.43 $\pm$ 0.74&
84.05 $\pm$ 0.45&
39.59 $\pm$ 0.38&
64.73 $\pm$ 0.31 \\
20W-5S&
Inception without Pooling&
70.11 $\pm$ 0.76&
\textbf{85.65 $\pm$ 0.45}&
\textbf{45.14 $\pm$ 0.38}&
66.25 $\pm$ 0.30 \\
20W-5S&
Inception Module&
\textbf{70.40 $\pm$ 0.72}&
84.95 $\pm$ 0.47&
44.59 $\pm$ 0.37&
65.66 $\pm$ 0.30 \\
20W-5S&
\emph{Sum12} Multi-Scale with Pooling&
66.67 $\pm$ 0.73&
84.79 $\pm$ 0.44&
40.89 $\pm$ 0.36&
\textbf{66.35 $\pm$ 0.29} \\
\hline
\end{tabular}
\end{table*}

\begin{table*}[!htbp]
\caption{\textbf{Few-shot classification accuracy (\%) on Omniglot under different fusion strategies. Best results per training and evaluation configuration are highlighted in \textbf{bold}.}}
\label{tab:omniglot_fusion_strategies}
\centering
\small
\setlength{\tabcolsep}{2pt}
\begin{tabular}{|p{45pt}|p{125pt}|p{70pt}|p{70pt}|p{70pt}|p{70pt}|}
\hline
Trained at&
Fusion Type&
5W-1S&
5W-5S&
20W-1S&
20W-5S \\
\hline
5W-1S&
Uniform Avg across branches&
97.80 $\pm$ 0.25&
99.43 $\pm$ 0.12&
93.72 $\pm$ 0.21&
98.03 $\pm$ 0.11 \\
5W-1S&
Scalar per-branch weights&
97.55 $\pm$ 0.26&
99.37 $\pm$ 0.11&
92.67 $\pm$ 0.23&
97.68 $\pm$ 0.14 \\
5W-1S&
SE-like gating&
97.86 $\pm$ 0.25&
99.33 $\pm$ 0.12&
93.76 $\pm$ 0.22&
97.99 $\pm$ 0.12 \\
5W-1S&
Weighted Average&
97.85 $\pm$ 0.25&
99.41 $\pm$ 0.10&
93.76 $\pm$ 0.21&
98.10 $\pm$ 0.11 \\
5W-1S&
Concat + \(1\times1\) conv&
97.59 $\pm$ 0.25&
99.28 $\pm$ 0.13&
92.69 $\pm$ 0.23&
97.69 $\pm$ 0.13 \\
5W-1S&
Attention&
\textbf{97.97 $\pm$ 0.24}&
\textbf{99.46 $\pm$ 0.10}&
\textbf{93.79 $\pm$ 0.20}&
\textbf{98.15 $\pm$ 0.11} \\
5W-1S&
\(1\times1\) conv&
97.63 $\pm$ 0.25&
99.31 $\pm$ 0.12&
92.97 $\pm$ 0.22&
98.00 $\pm$ 0.11 \\
\hline
5W-5S&
Uniform Avg across branches&
97.01 $\pm$ 0.30&
99.42 $\pm$ 0.10&
92.42 $\pm$ 0.21&
\textbf{98.23 $\pm$ 0.09} \\
5W-5S&
Scalar per-branch weights&
97.09 $\pm$ 0.28&
99.38 $\pm$ 0.10&
92.43 $\pm$ 0.21&
98.10 $\pm$ 0.10 \\
5W-5S&
SE-like gating&
\textbf{97.64 $\pm$ 0.25}&
\textbf{99.45 $\pm$ 0.11}&
\textbf{92.86 $\pm$ 0.22}&
\textbf{98.23 $\pm$ 0.10} \\
5W-5S&
Weighted Average&
96.80 $\pm$ 0.29&
99.28 $\pm$ 0.12&
91.99 $\pm$ 0.23&
97.80 $\pm$ 0.11 \\
5W-5S&
Concat + \(1\times1\) conv&
96.32 $\pm$ 0.32&
99.25 $\pm$ 0.12&
91.02 $\pm$ 0.24&
97.61 $\pm$ 0.12 \\
5W-5S&
Attention&
97.07 $\pm$ 0.27&
99.32 $\pm$ 0.10&
92.02 $\pm$ 0.23&
97.89 $\pm$ 0.11 \\
5W-5S&
\(1\times1\) conv&
96.92 $\pm$ 0.26&
99.39 $\pm$ 0.08&
90.32 $\pm$ 0.24&
97.77 $\pm$ 0.10 \\
\hline
20W-1S&
Uniform Avg across branches&
98.90 $\pm$ 0.18&
\textbf{99.78 $\pm$ 0.05}&
\textbf{96.51 $\pm$ 0.17}&
99.04 $\pm$ 0.08 \\
20W-1S&
Scalar per-branch weights&
\textbf{99.10 $\pm$ 0.15}&
99.68 $\pm$ 0.08&
96.42 $\pm$ 0.17&
98.98 $\pm$ 0.08 \\
20W-1S&
SE-like gating&
98.80 $\pm$ 0.20&
99.72 $\pm$ 0.08&
96.34 $\pm$ 0.18&
99.05 $\pm$ 0.08 \\
20W-1S&
Weighted Average&
98.87 $\pm$ 0.20&
\textbf{99.78 $\pm$ 0.05}&
96.17 $\pm$ 0.18&
98.98 $\pm$ 0.09 \\
20W-1S&
Concat + \(1\times1\) conv&
98.71 $\pm$ 0.21&
99.75 $\pm$ 0.07&
96.22 $\pm$ 0.17&
98.93 $\pm$ 0.09 \\
20W-1S&
Attention&
98.96 $\pm$ 0.19&
99.74 $\pm$ 0.06&
96.39 $\pm$ 0.18&
\textbf{99.09 $\pm$ 0.08} \\
20W-1S&
\(1\times1\) conv&
98.90 $\pm$ 0.18&
99.76 $\pm$ 0.06&
96.21 $\pm$ 0.17&
99.03 $\pm$ 0.07 \\
\hline
20W-5S&
Uniform Avg across branches&
98.62 $\pm$ 0.20&
99.74 $\pm$ 0.06&
95.34 $\pm$ 0.20&
98.95 $\pm$ 0.08 \\
20W-5S&
Scalar per-branch weights&
98.65 $\pm$ 0.19&
99.73 $\pm$ 0.08&
95.19 $\pm$ 0.19&
98.98 $\pm$ 0.08 \\
20W-5S&
SE-like gating&
\textbf{98.69 $\pm$ 0.20}&
99.73 $\pm$ 0.10&
\textbf{95.76 $\pm$ 0.18}&
\textbf{99.17 $\pm$ 0.07} \\
20W-5S&
Weighted Average&
98.50 $\pm$ 0.21&
\textbf{99.78 $\pm$ 0.06}&
95.30 $\pm$ 0.19&
98.99 $\pm$ 0.09 \\
20W-5S&
Concat + \(1\times1\) conv&
98.11 $\pm$ 0.25&
99.69 $\pm$ 0.09&
94.67 $\pm$ 0.20&
98.85 $\pm$ 0.09 \\
20W-5S&
Attention&
98.48 $\pm$ 0.20&
99.68 $\pm$ 0.09&
95.50 $\pm$ 0.18&
98.93 $\pm$ 0.08 \\
20W-5S&
\(1\times1\) conv&
98.26 $\pm$ 0.22&
99.74 $\pm$ 0.07&
94.92 $\pm$ 0.19&
99.03 $\pm$ 0.07 \\
\hline
\end{tabular}
\end{table*}

\begin{table*}[!htbp]
\caption{\textbf{Few-shot classification accuracy (\%) on CIFAR-FS under different fusion strategies. Best results per training and evaluation configuration are highlighted in \textbf{bold}.}}
\label{tab:cifarfs_fusion_strategies}
\centering
\small
\setlength{\tabcolsep}{2pt}
\begin{tabular}{|p{45pt}|p{125pt}|p{70pt}|p{70pt}|p{70pt}|p{70pt}|}
\hline
Trained at&
Fusion Type&
5W-1S&
5W-5S&
20W-1S&
20W-5S \\
\hline
5W-1S&
Uniform Avg across branches&
63.61 $\pm$ 0.78&
77.77 $\pm$ 0.56&
35.39 $\pm$ 0.33&
53.07 $\pm$ 0.30 \\
5W-1S&
Scalar per-branch weights&
62.26 $\pm$ 0.76&
77.04 $\pm$ 0.57&
35.13 $\pm$ 0.34&
52.61 $\pm$ 0.30 \\
5W-1S&
SE-like gating&
\textbf{65.73 $\pm$ 0.73}&
\textbf{78.99 $\pm$ 0.55}&
\textbf{37.49 $\pm$ 0.33}&
\textbf{55.14 $\pm$ 0.30} \\
5W-1S&
Weighted Average&
63.65 $\pm$ 0.78&
77.85 $\pm$ 0.56&
35.38 $\pm$ 0.32&
53.01 $\pm$ 0.31 \\
5W-1S&
Concat + \(1\times1\) conv&
61.01 $\pm$ 0.77&
75.97 $\pm$ 0.57&
34.01 $\pm$ 0.33&
51.52 $\pm$ 0.30 \\
5W-1S&
Attention&
64.29 $\pm$ 0.75&
78.07 $\pm$ 0.55&
36.15 $\pm$ 0.33&
53.78 $\pm$ 0.30 \\
5W-1S&
\(1\times1\) conv&
60.69 $\pm$ 0.77&
75.37 $\pm$ 0.57&
32.98 $\pm$ 0.31&
50.90 $\pm$ 0.30 \\
\hline
5W-5S&
Uniform Avg across branches&
63.80 $\pm$ 0.76&
81.69 $\pm$ 0.51&
36.08 $\pm$ 0.34&
59.16 $\pm$ 0.30 \\
5W-5S&
Scalar per-branch weights&
63.35 $\pm$ 0.70&
81.46 $\pm$ 0.50&
36.04 $\pm$ 0.32&
59.35 $\pm$ 0.30 \\
5W-5S&
SE-like gating&
\textbf{66.32 $\pm$ 0.74}&
\textbf{82.67 $\pm$ 0.50}&
\textbf{38.91 $\pm$ 0.33}&
\textbf{60.83 $\pm$ 0.31} \\
5W-5S&
Weighted Average&
64.91 $\pm$ 0.74&
82.07 $\pm$ 0.50&
37.07 $\pm$ 0.35&
59.97 $\pm$ 0.30 \\
5W-5S&
Concat + \(1\times1\) conv&
63.34 $\pm$ 0.73&
80.85 $\pm$ 0.52&
35.82 $\pm$ 0.33&
58.65 $\pm$ 0.29 \\
5W-5S&
Attention&
65.03 $\pm$ 0.72&
81.87 $\pm$ 0.50&
37.51 $\pm$ 0.32&
59.36 $\pm$ 0.30 \\
5W-5S&
\(1\times1\) conv&
62.60 $\pm$ 0.74&
80.78 $\pm$ 0.51&
35.69 $\pm$ 0.34&
58.74 $\pm$ 0.31 \\
\hline
20W-1S&
Uniform Avg across branches&
71.53 $\pm$ 0.77&
83.44 $\pm$ 0.51&
45.32 $\pm$ 0.39&
62.91 $\pm$ 0.32 \\
20W-1S&
Scalar per-branch weights&
71.48 $\pm$ 0.75&
83.69 $\pm$ 0.49&
45.79 $\pm$ 0.38&
63.47 $\pm$ 0.32 \\
20W-1S&
SE-like gating&
\textbf{73.75 $\pm$ 0.74}&
\textbf{84.91 $\pm$ 0.49}&
\textbf{48.40 $\pm$ 0.40}&
\textbf{66.37 $\pm$ 0.31} \\
20W-1S&
Weighted Average&
71.77 $\pm$ 0.78&
83.57 $\pm$ 0.50&
45.72 $\pm$ 0.39&
63.42 $\pm$ 0.33 \\
20W-1S&
Concat + \(1\times1\) conv&
70.30 $\pm$ 0.74&
82.84 $\pm$ 0.50&
44.18 $\pm$ 0.38&
61.82 $\pm$ 0.33 \\
20W-1S&
Attention&
71.20 $\pm$ 0.75&
83.06 $\pm$ 0.51&
45.28 $\pm$ 0.37&
63.51 $\pm$ 0.32 \\
20W-1S&
\(1\times1\) conv&
70.76 $\pm$ 0.77&
82.37 $\pm$ 0.52&
44.01 $\pm$ 0.39&
61.80 $\pm$ 0.32 \\
\hline
20W-5S&
Uniform Avg across branches&
62.73 $\pm$ 0.73&
83.53 $\pm$ 0.48&
37.67 $\pm$ 0.36&
63.83 $\pm$ 0.30 \\
20W-5S&
Scalar per-branch weights&
64.91 $\pm$ 0.74&
83.99 $\pm$ 0.45&
39.16 $\pm$ 0.37&
64.69 $\pm$ 0.30 \\
20W-5S&
SE-like gating&
\textbf{66.74 $\pm$ 0.76}&
\textbf{85.05 $\pm$ 0.47}&
\textbf{41.84 $\pm$ 0.37}&
\textbf{66.78 $\pm$ 0.28} \\
20W-5S&
Weighted Average&
64.08 $\pm$ 0.76&
84.06 $\pm$ 0.47&
38.21 $\pm$ 0.36&
64.50 $\pm$ 0.29 \\
20W-5S&
Concat + \(1\times1\) conv&
62.56 $\pm$ 0.73&
82.12 $\pm$ 0.49&
36.39 $\pm$ 0.35&
61.40 $\pm$ 0.30 \\
20W-5S&
Attention&
64.10 $\pm$ 0.75&
84.21 $\pm$ 0.48&
38.90 $\pm$ 0.36&
64.70 $\pm$ 0.29 \\
20W-5S&
\(1\times1\) conv&
62.98 $\pm$ 0.74&
82.13 $\pm$ 0.50&
37.27 $\pm$ 0.33&
61.78 $\pm$ 0.31 \\
\hline
\end{tabular}
\end{table*}

\begin{table*}[!htbp]
\caption{\textbf{Classification accuracy (\%) on CIFAR-FS with different encoder types. Best results per training and evaluation configuration are highlighted in \textbf{bold}.}}
\label{tab:varying_sum_ablation}
\centering
\small
\setlength{\tabcolsep}{2pt}
\begin{tabular}{|p{45pt}|p{55pt}|p{70pt}|p{70pt}|p{70pt}|p{70pt}|}
\hline
Trained at&
Encoder Type&
5W-1S&
5W-5S&
20W-1S&
20W-5S \\
\hline
5W-1S&
\emph{Sum3}&
43.03 $\pm$ 0.66&
58.10 $\pm$ 0.64&
18.98 $\pm$ 0.23&
31.65 $\pm$ 0.26 \\
5W-1S&
\emph{Sum6}&
54.37 $\pm$ 0.74&
70.42 $\pm$ 0.60&
27.74 $\pm$ 0.30&
44.38 $\pm$ 0.29 \\
5W-1S&
\emph{Sum9}&
60.27 $\pm$ 0.75&
75.39 $\pm$ 0.56&
32.76 $\pm$ 0.31&
50.25 $\pm$ 0.30 \\
5W-1S&
\emph{Sum12}&
\textbf{63.13 $\pm$ 0.76}&
77.61 $\pm$ 0.57&
35.10 $\pm$ 0.33&
53.09 $\pm$ 0.32 \\
5W-1S&
\emph{Sum15}&
62.71 $\pm$ 0.78&
77.56 $\pm$ 0.55&
\textbf{35.85 $\pm$ 0.34}&
\textbf{53.83 $\pm$ 0.31} \\
5W-1S&
\emph{Sum18}&
\hl{62.58 $\pm$ 0.77}&
\textbf{77.86 $\pm$ 0.54}&
35.62 $\pm$ 0.34&
53.14 $\pm$ 0.31 \\
\hline
5W-5S&
\emph{Sum3}&
46.97 $\pm$ 0.68&
70.17 $\pm$ 0.57&
23.02 $\pm$ 0.27&
44.05 $\pm$ 0.29 \\
5W-5S&
\emph{Sum6}&
55.31 $\pm$ 0.76&
77.20 $\pm$ 0.55&
29.45 $\pm$ 0.32&
52.64 $\pm$ 0.30 \\
5W-5S&
\emph{Sum9}&
62.40 $\pm$ 0.72&
80.38 $\pm$ 0.54&
34.87 $\pm$ 0.32&
58.16 $\pm$ 0.30 \\
5W-5S&
\emph{Sum12}&
64.49 $\pm$ 0.73&
82.12 $\pm$ 0.50&
37.05 $\pm$ 0.33&
60.10 $\pm$ 0.30 \\
5W-5S&
\emph{Sum15}&
65.60 $\pm$ 0.71&
82.03 $\pm$ 0.51&
38.08 $\pm$ 0.34&
60.22 $\pm$ 0.30 \\
5W-5S&
\emph{Sum18}&
\textbf{66.35 $\pm$ 0.74}&
\textbf{82.71 $\pm$ 0.49}&
\textbf{39.35 $\pm$ 0.34}&
\textbf{61.49 $\pm$ 0.29} \\
\hline
20W-1S&
\emph{Sum3}&
48.43 $\pm$ 0.74&
63.89 $\pm$ 0.67&
22.71 $\pm$ 0.26&
37.31 $\pm$ 0.26 \\
20W-1S&
\emph{Sum6}&
\hl{62.57 $\pm$ 0.78}&
77.51 $\pm$ 0.57&
35.40 $\pm$ 0.35&
52.98 $\pm$ 0.31 \\
20W-1S&
\emph{Sum9}&
68.02 $\pm$ 0.78&
82.53 $\pm$ 0.50&
41.50 $\pm$ 0.39&
60.26 $\pm$ 0.33 \\
20W-1S&
\emph{Sum12}&
71.66 $\pm$ 0.78&
83.84 $\pm$ 0.51&
45.99 $\pm$ 0.38&
63.59 $\pm$ 0.32 \\
20W-1S&
\emph{Sum15}&
72.79 $\pm$ 0.79&
84.66 $\pm$ 0.49&
47.50 $\pm$ 0.40&
64.82 $\pm$ 0.32 \\
20W-1S&
\emph{Sum18}&
\textbf{73.57 $\pm$ 0.74}&
\textbf{85.19 $\pm$ 0.47}&
\textbf{49.35 $\pm$ 0.39}&
\textbf{66.30 $\pm$ 0.32} \\
\hline
20W-5S&
\emph{Sum3}&
50.88 $\pm$ 0.72&
73.63 $\pm$ 0.61&
25.33 $\pm$ 0.29&
49.37 $\pm$ 0.31 \\
20W-5S&
\emph{Sum6}&
60.06 $\pm$ 0.78&
82.18 $\pm$ 0.49&
33.80 $\pm$ 0.35&
60.94 $\pm$ 0.31 \\
20W-5S&
\emph{Sum9}&
61.96 $\pm$ 0.75&
83.33 $\pm$ 0.48&
35.78 $\pm$ 0.35&
62.85 $\pm$ 0.30 \\
20W-5S&
\emph{Sum12}&
64.32 $\pm$ 0.75&
83.67 $\pm$ 0.47&
38.00 $\pm$ 0.35&
64.14 $\pm$ 0.29 \\
20W-5S&
\emph{Sum15}&
66.75 $\pm$ 0.75&
84.55 $\pm$ 0.47&
42.23 $\pm$ 0.38&
65.70 $\pm$ 0.30 \\
20W-5S&
\emph{Sum18}&
\textbf{68.52 $\pm$ 0.73}&
\textbf{84.71 $\pm$ 0.45}&
\textbf{43.65 $\pm$ 0.38}&
\textbf{66.58 $\pm$ 0.30} \\
\hline
\end{tabular}
\end{table*}

\begin{figure*}[!b]
    \centering
    \includegraphics[width=1\textwidth,height=0.8\textheight,keepaspectratio]{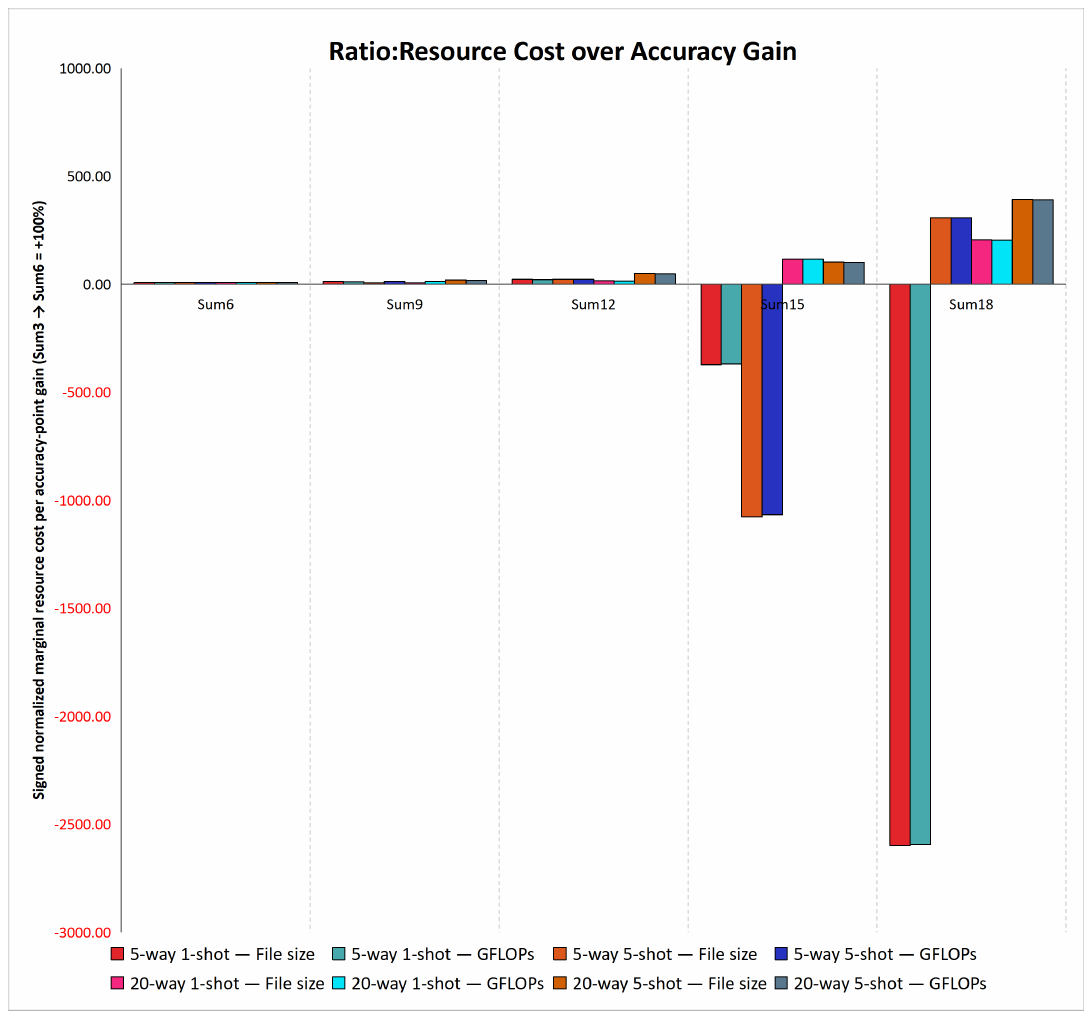}
    \caption{\hl{Resource cost per accuracy gain for varying target sums}.}
    \label{fig:resourcecost_accuracygain}
\end{figure*}

\begin{table*}[!htbp]
\caption{\textbf{Branch-level analysis on CIFAR-FS. Classification accuracy (\%) is reported for each individual branch, with the corresponding average shown for comparison. Italicized entries indicate branches identified as less effective in the corresponding analysis.}}
\label{tab:cifarfs_branches_individually}
\centering
\small
\setlength{\tabcolsep}{2pt}
\begin{tabular}{|p{38pt}|p{35pt}|p{60pt}|p{25pt}|p{60pt}|p{25pt}|p{60pt}|p{25pt}|p{60pt}|p{25pt}|}
\hline
Trained at&
Branch&
5W-1S&
AVG&
5W-5S&
AVG&
20W-1S&
AVG&
20W-5S&
AVG \\
\hline
5W-1S&
3333&
54.29 $\pm$ 0.76&
53.6&
68.56 $\pm$ 0.62&
69.4&
26.79 $\pm$ 0.28&
26.8&
42.25 $\pm$ 0.28&
43.1 \\
5W-1S&
336&
54.57 $\pm$ 0.76&
53.6&
70.17 $\pm$ 0.59&
69.4&
27.22 $\pm$ 0.28&
26.8&
43.73 $\pm$ 0.28&
43.1 \\
5W-1S&
363&
54.25 $\pm$ 0.74&
53.6&
70.66 $\pm$ 0.60&
69.4&
27.25 $\pm$ 0.29&
26.8&
43.82 $\pm$ 0.28&
43.1 \\
5W-1S&
633&
\textit{52.52 $\pm$ 0.73}&
53.6&
\textit{68.10 $\pm$ 0.63}&
69.4&
\textit{25.69 $\pm$ 0.28}&
26.8&
\textit{42.01 $\pm$ 0.29}&
43.1 \\
5W-1S&
66&
\textit{53.09 $\pm$ 0.73}&
53.6&
\textit{68.58 $\pm$ 0.62}&
69.4&
\textit{26.36 $\pm$ 0.28}&
26.8&
\textit{42.71 $\pm$ 0.27}&
43.1 \\
5W-1S&
93&
\textit{52.16 $\pm$ 0.74}&
53.6&
\textit{68.86 $\pm$ 0.68}&
69.4&
\textit{26.24 $\pm$ 0.28}&
26.8&
\textit{42.54 $\pm$ 0.28}&
43.1 \\
5W-1S&
39&
55.11 $\pm$ 0.73&
53.6&
71.59 $\pm$ 0.60&
69.4&
28.29 $\pm$ 0.29&
26.8&
45.19 $\pm$ 0.28&
43.1 \\
\hline
5W-5S&
3333&
57.42 $\pm$ 0.73&
56.6&
76.61 $\pm$ 0.56&
76.3&
31.19 $\pm$ 0.29&
30.2&
51.56 $\pm$ 0.30&
51.7 \\
5W-5S&
336&
57.52 $\pm$ 0.73&
56.6&
77.35 $\pm$ 0.53&
76.3&
31.26 $\pm$ 0.28&
30.2&
52.87 $\pm$ 0.31&
51.7 \\
5W-5S&
363&
57.67 $\pm$ 0.74&
56.6&
76.77 $\pm$ 0.54&
76.3&
30.77 $\pm$ 0.32&
30.2&
52.38 $\pm$ 0.31&
51.7 \\
5W-5S&
633&
\textit{56.17 $\pm$ 0.71}&
56.6&
\textit{76.35 $\pm$ 0.55}&
76.3&
\textit{29.89 $\pm$ 0.30}&
30.2&
\textit{50.99 $\pm$ 0.29}&
51.7 \\
5W-5S&
66&
\textit{55.75 $\pm$ 0.71}&
56.6&
\textit{75.25 $\pm$ 0.53}&
76.3&
\textit{29.28 $\pm$ 0.30}&
30.2&
\textit{51.01 $\pm$ 0.30}&
51.7 \\
5W-5S&
93&
\textit{56.06 $\pm$ 0.78}&
56.6&
\textit{75.84 $\pm$ 0.55}&
76.3&
\textit{29.34 $\pm$ 0.31}&
30.2&
\textit{50.73 $\pm$ 0.30}&
51.7 \\
5W-5S&
39&
56.65 $\pm$ 0.72&
56.6&
76.95 $\pm$ 0.55&
76.3&
30.09 $\pm$ 0.30&
30.2&
52.33 $\pm$ 0.31&
51.7 \\
\hline
20W-1S&
3333&
63.29 $\pm$ 0.79&
61.7&
76.75 $\pm$ 0.58&
76.3&
35.35 $\pm$ 0.35&
34.1&
51.96 $\pm$ 0.30&
51.7 \\
20W-1S&
336&
\textit{61.47 $\pm$ 0.79}&
61.7&
77.20 $\pm$ 0.56&
76.3&
\textit{34.10 $\pm$ 0.34}&
34.1&
52.24 $\pm$ 0.30&
51.7 \\
20W-1S&
363&
62.64 $\pm$ 0.79&
61.7&
76.32 $\pm$ 0.58&
76.3&
34.51 $\pm$ 0.33&
34.1&
52.03 $\pm$ 0.31&
51.7 \\
20W-1S&
633&
\textit{61.99 $\pm$ 0.76}&
61.7&
\textit{76.27 $\pm$ 0.56}&
76.3&
\textit{34.69 $\pm$ 0.34}&
34.1&
\textit{51.54 $\pm$ 0.31}&
51.7 \\
20W-1S&
66&
\textit{60.15 $\pm$ 0.78}&
61.7&
\textit{75.14 $\pm$ 0.58}&
76.3&
\textit{32.48 $\pm$ 0.32}&
34.1&
\textit{50.39 $\pm$ 0.30}&
51.7 \\
20W-1S&
93&
\textit{59.92 $\pm$ 0.76}&
61.7&
\textit{75.78 $\pm$ 0.57}&
76.3&
\textit{33.22 $\pm$ 0.32}&
34.1&
\textit{51.22 $\pm$ 0.31}&
51.7 \\
20W-1S&
39&
62.95 $\pm$ 0.78&
61.7&
77.08 $\pm$ 0.56&
76.3&
34.52 $\pm$ 0.33&
34.1&
53.14 $\pm$ 0.30&
51.7 \\
\hline
20W-5S&
3333&
61.56 $\pm$ 0.75&
60.3&
81.65 $\pm$ 0.56&
80.8&
35.53 $\pm$ 0.34&
34.0&
58.39 $\pm$ 0.31&
58.4 \\
20W-5S&
336&
60.97 $\pm$ 0.75&
60.3&
81.28 $\pm$ 0.51&
80.8&
35.16 $\pm$ 0.33&
34.0&
59.45 $\pm$ 0.31&
58.4 \\
20W-5S&
363&
61.77 $\pm$ 0.76&
60.3&
81.68 $\pm$ 0.53&
80.8&
34.63 $\pm$ 0.33&
34.0&
59.41 $\pm$ 0.31&
58.4 \\
20W-5S&
633&
\textit{60.35 $\pm$ 0.72}&
60.3&
\textit{80.68 $\pm$ 0.52}&
80.8&
\textit{33.66 $\pm$ 0.34}&
34.0&
\textit{57.03 $\pm$ 0.31}&
58.4 \\
20W-5S&
66&
\textit{60.16 $\pm$ 0.74}&
60.3&
\textit{80.60 $\pm$ 0.52}&
80.8&
\textit{33.75 $\pm$ 0.34}&
34.0&
\textit{58.74 $\pm$ 0.31}&
58.4 \\
20W-5S&
93&
\textit{58.33 $\pm$ 0.75}&
60.3&
\textit{80.57 $\pm$ 0.49}&
80.8&
\textit{32.31 $\pm$ 0.33}&
34.0&
\textit{57.40 $\pm$ 0.30}&
58.4 \\
20W-5S&
39&
60.75 $\pm$ 0.75&
60.3&
81.65 $\pm$ 0.51&
80.8&
33.03 $\pm$ 0.33&
34.0&
58.67 $\pm$ 0.31&
58.4 \\
\hline
\end{tabular}
\end{table*}

\clearpage          

\section{Extended MedMNIST 2D Results}\label{app:medmnist}

\begin{table*}[!b]
\caption{\textbf{Comprehensive results on MedMNIST v2 in terms of AUC and ACC. Notes: Values are mean AUC/ACC over three training runs. ``(28)'' and ``(224)'' denote input resolutions of $28{\times}28$ and $224{\times}224$, respectively. $^{*}$ For BreastMNIST at $28{\times}28$, we report the midpoint of the performance range reported in \mbox{Table~\ref{tab:breastmnist}}.}}
\label{tab:medmnist_all_panels}
\centering
\small
\setlength{\tabcolsep}{2pt}
\begin{tabular}{|p{135pt}|p{45pt}|p{45pt}|p{45pt}|p{45pt}|p{45pt}|p{45pt}|}
\hline
Method&
\shortstack{Path\\AUC}&
\shortstack{Path\\ACC}&
\shortstack{Chest\\AUC}&
\shortstack{Chest\\ACC}&
\shortstack{Derma\\AUC}&
\shortstack{Derma\\ACC} \\
\hline
ResNet-18 (28) \cite{medmnist,resnetCitation}&
0.983&0.907&0.768&\textit{0.947}&0.917&0.735 \\
ResNet-18 (224) \cite{medmnist,resnetCitation}&
\emph{0.989}&0.909&\emph{0.773}&\textit{0.947}&0.920&0.758 \\
ResNet-50 (28) \cite{medmnist,resnetCitation}&
\textbf{0.990}&\emph{0.911}&0.769&\textit{0.947}&0.913&0.735 \\
ResNet-50 (224) \cite{medmnist,resnetCitation}&
\emph{0.989}&0.892&\emph{0.773}&\textbf{0.948}&0.912&0.731 \\
Auto-sklearn \cite{medmnist,autosklearnCitation}&
0.934&0.716&0.649&0.779&0.902&0.719 \\
AutoKeras \cite{medmnist,autokerasCitation}&
0.959&0.834&0.742&0.937&0.915&0.749 \\
Google AutoML Vision \cite{medmnist}&
0.944&0.728&\textbf{0.778}&\textbf{0.948}&0.914&\emph{0.768} \\
\emph{Sum12} Multi-Scale (28)&
0.980&0.889&0.704&\textbf{0.948}&\textbf{0.931}&\textbf{0.780} \\
\emph{Sum12} Multi-Scale (224)&
0.985&\textbf{0.912}&0.635&\textit{0.947}&\emph{0.930}&0.758 \\
\hline
Method&
\shortstack{OCT\\AUC}&
\shortstack{OCT\\ACC}&
\shortstack{Pneum\\AUC}&
\shortstack{Pneum\\ACC}&
\shortstack{Retina\\AUC}&
\shortstack{Retina\\ACC} \\
\hline
ResNet-18 (28) \cite{medmnist,resnetCitation}&
0.943&0.743&0.944&0.854&0.717&0.524 \\
ResNet-18 (224) \cite{medmnist,resnetCitation}&
0.958&0.763&0.956&0.864&0.710&0.493 \\
ResNet-50 (28) \cite{medmnist,resnetCitation}&
0.952&0.762&0.948&0.854&0.726&\emph{0.528} \\
ResNet-50 (224) \cite{medmnist,resnetCitation}&
0.958&0.776&0.962&0.884&0.716&0.511 \\
Auto-sklearn \cite{medmnist,autosklearnCitation}&
0.887&0.601&0.942&0.855&0.690&0.515 \\
AutoKeras \cite{medmnist,autokerasCitation}&
0.955&0.763&0.947&0.878&0.719&0.503 \\
Google AutoML Vision \cite{medmnist}&
\emph{0.963}&0.771&\textbf{0.991}&\textbf{0.946}&\emph{0.750}&\textbf{0.531} \\
\emph{Sum12} Multi-Scale (28)&
0.955&\emph{0.786}&\emph{0.975}&0.915&0.733&0.525 \\
\emph{Sum12} Multi-Scale (224)&
\textbf{0.969}&\textbf{0.795}&0.971&\emph{0.918}&0.729&\textbf{0.531} \\
\hline
Method&
\shortstack{Breast\\AUC}&
\shortstack{Breast\\ACC}&
\shortstack{Blood\\AUC}&
\shortstack{Blood\\ACC}&
\shortstack{Tissue\\AUC}&
\shortstack{Tissue\\ACC} \\
\hline
ResNet-18 (28) \cite{medmnist,resnetCitation}&
\textit{0.901}&\textbf{0.863}&0.998&0.958&0.930&0.676 \\
ResNet-18 (224) \cite{medmnist,resnetCitation}&
0.891&0.833&0.998&0.963&\emph{0.933}&\emph{0.681} \\
ResNet-50 (28) \cite{medmnist,resnetCitation}&
0.857&0.812&0.997&0.956&0.931&0.680 \\
ResNet-50 (224) \cite{medmnist,resnetCitation}&
0.866&0.842&0.997&0.950&0.932&0.680 \\
Auto-sklearn \cite{medmnist,autosklearnCitation}&
0.836&0.803&0.984&0.878&0.828&0.532 \\
AutoKeras \cite{medmnist,autokerasCitation}&
0.871&0.831&0.998&0.961&\textbf{0.941}&\textbf{0.703} \\
Google AutoML Vision \cite{medmnist}&
\textbf{0.919}&\emph{0.861}&0.998&0.966&0.924&0.673 \\
\emph{Sum12} Multi-Scale (28)$^{*}$&
0.896&0.820&0.998&0.965&0.930&\emph{0.681} \\
\emph{Sum12} Multi-Scale (224)&
--&--&\textbf{0.999}&\textbf{0.973}&0.926&0.670 \\
\hline
Method&
\shortstack{OrganA\\AUC}&
\shortstack{OrganA\\ACC}&
\shortstack{OrganC\\AUC}&
\shortstack{OrganC\\ACC}&
\shortstack{OrganS\\AUC}&
\shortstack{OrganS\\ACC} \\
\hline
ResNet-18 (28) \cite{medmnist,resnetCitation}&
\emph{0.997}&0.935&0.992&0.900&0.972&0.782 \\
ResNet-18 (224) \cite{medmnist,resnetCitation}&
\textbf{0.998}&\textbf{0.951}&\emph{0.994}&\emph{0.920}&0.974&0.778 \\
ResNet-50 (28) \cite{medmnist,resnetCitation}&
\emph{0.997}&0.935&0.992&0.905&0.972&0.770 \\
ResNet-50 (224) \cite{medmnist,resnetCitation}&
\textbf{0.998}&\emph{0.947}&0.993&0.911&\emph{0.975}&0.785 \\
Auto-sklearn \cite{medmnist,autosklearnCitation}&
0.963&0.762&0.976&0.829&0.945&0.672 \\
AutoKeras \cite{medmnist,autokerasCitation}&
0.994&0.905&0.990&0.879&0.974&\textbf{0.813} \\
Google AutoML Vision \cite{medmnist}&
0.990&0.886&0.988&0.877&0.964&0.749 \\
\emph{Sum12} Multi-Scale (28)&
\emph{0.997}&0.939&\textbf{0.995}&\textbf{0.928}&\emph{0.976}&\emph{0.810} \\
\emph{Sum12} Multi-Scale (224)&
\emph{0.997}&\emph{0.943}&0.993&0.918&\textbf{0.980}&\emph{0.810} \\
\hline
\end{tabular}
\end{table*}

\clearpage

\section{\hl{STM32 Deployment Images}}\label{app:stm32deployment}
\begin{figure*}[!b]
    \centering
    \includegraphics[width=0.95\textwidth,height=0.72\textheight,keepaspectratio]{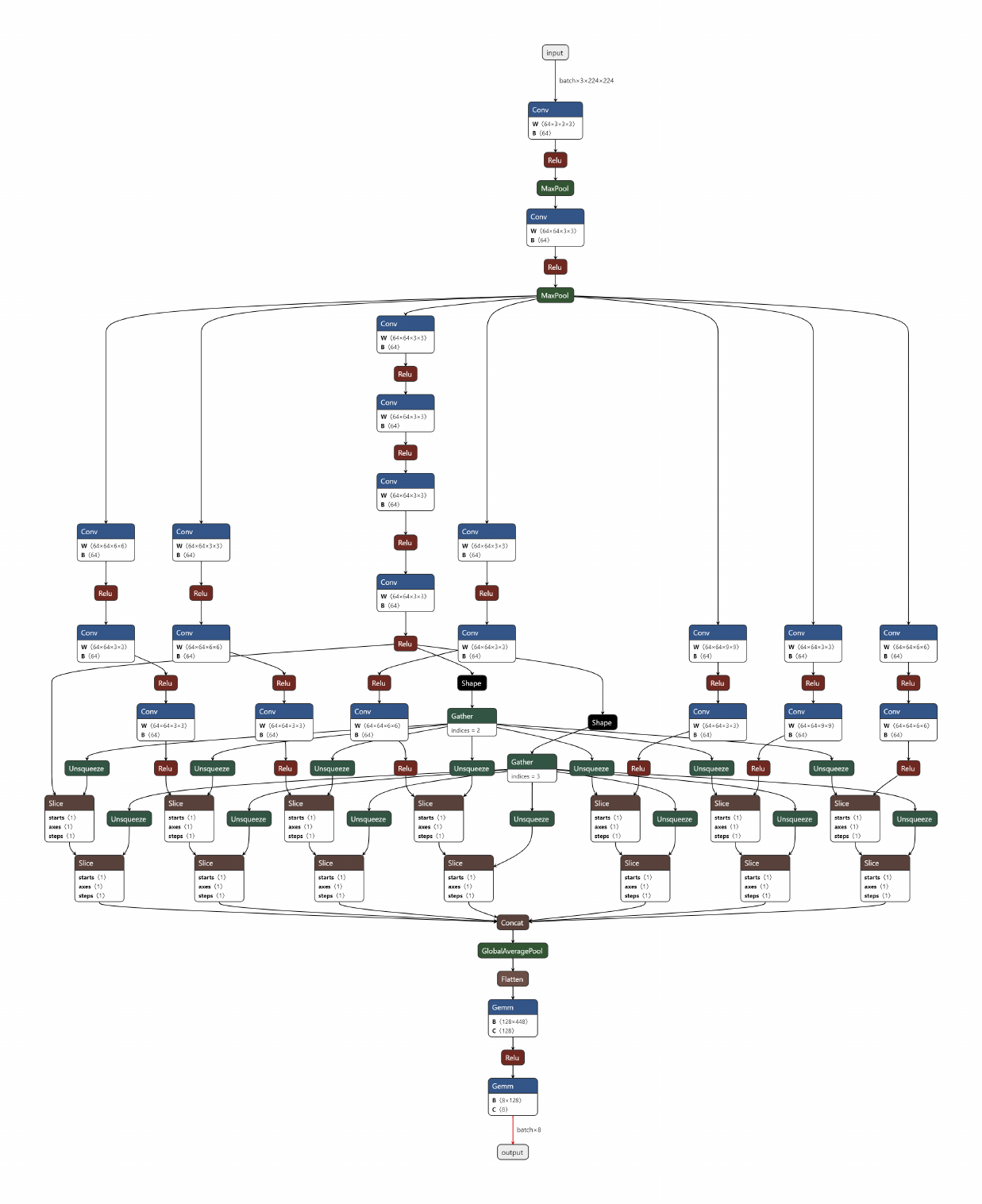}
    \caption{Graph produced by STM32CubeIDE showing the fully recognized model.}
    \label{fig:stm32graph_ide}
\end{figure*}

\begin{figure*}[!t]
    \centering

    \subfloat[\hl{Validation on host.}\label{fig:stm32_partial_quantization}]{
        \includegraphics[
            width=0.49\textwidth,
            trim=25mm 65mm 25mm 65mm,
            clip
        ]{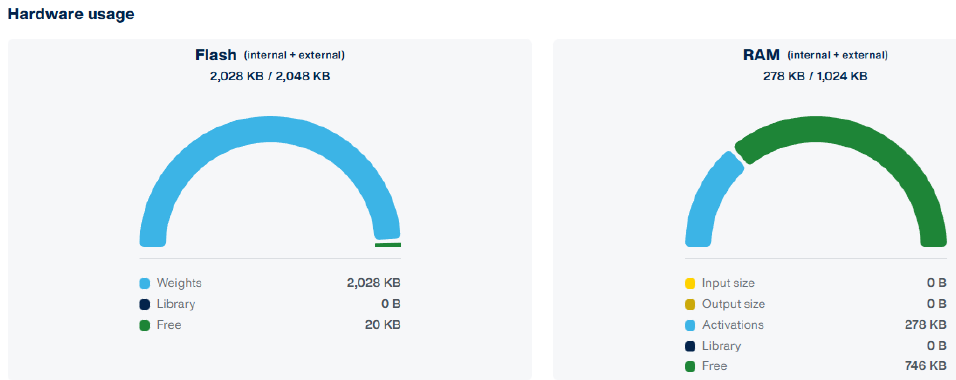}
    }
    \hfill
    \subfloat[\hl{Validation on the STM32 target.}\label{fig:stm32_on_target}]{
        \includegraphics[
            width=0.49\textwidth,
            trim=25mm 65mm 25mm 65mm,
            clip
        ]{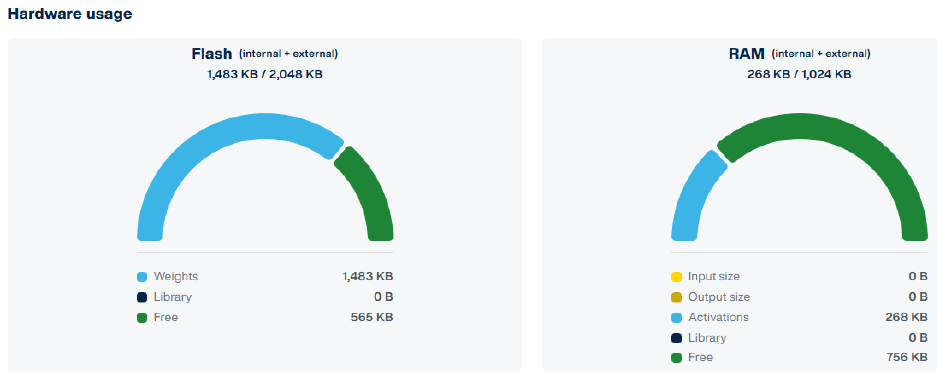}
    }

    \caption{\hl{STM32 AI Studio hardware-usage reports for the Sum12 encoder: (a) validation on host of the partially quantized model and (b) validation on
    the target STM32 platform of the fully INT8-quantized model.}}
    \label{fig:stm32_hardware_usage}
\end{figure*}

\begin{IEEEbiography}[{\includegraphics[width=1in,height=1.25in,clip,keepaspectratio]{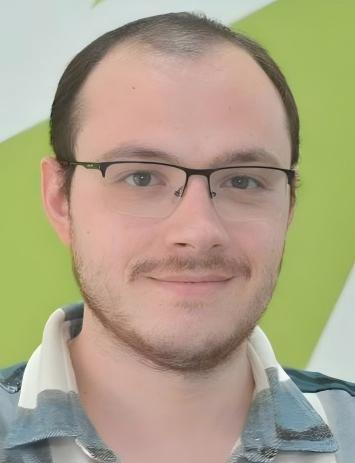}}]{Oriod Malo}
received the Bachelor's degree (Laurea Triennale) in Electronic and Telecommunication Engineering from Alma Mater Studiorum -- University of Bologna, Italy, in 2019, and the Master's degree (Laurea Magistrale) in Electronic Engineering from the same university in 2022. He is currently an Assistant Lecturer and Ph.D. candidate at University Metropolitan Tirana (UMT). His doctoral research focuses on machine learning, neural networks, and medical image analysis, with particular emphasis on the development and evaluation of artificial intelligence models for healthcare applications.

He received the Recherche Innovation France-Albanie (RIFA) scholarship in 2024 and 2025 for research periods at ESIEE Paris, Universit{\'e} Gustave Eiffel, France.
\end{IEEEbiography}

\begin{IEEEbiography}[{\includegraphics[width=1in,height=1.25in,clip,keepaspectratio]{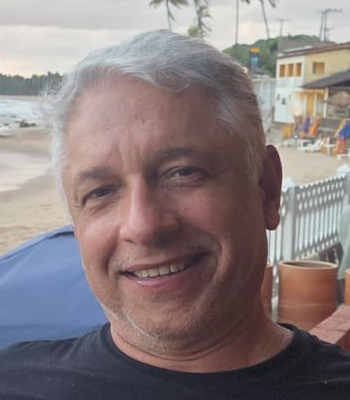}}]{Alexandre Xavier Falc{\~a}o} is a full professor at the Institute of
Computing (IC), University of Campinas (UNICAMP), with a B.Sc. (UFPE,
1984-1988), M.Sc. (UNICAMP, 1990-1993), and Ph.D. (UNICAMP, 1993-1996)
in Electrical Engineering. He is an IEEE Senior member, holds a CNPq
research productivity fellowship at level 1A, and currently serves as
Associate Editor at Pattern Recognition. He coordinates the Laboratory of
Image Data Science at UNICAMP and his research work aims at
computational models to learn, detect, delineate, identify, and
interpret the semantic content of images in the domain of several
applications. The areas of interest include image and video
processing, data visualization, medical image analysis, remote
sensing, graph algorithms, image annotation, organization, and
retrieval, and (deep) machine learning and pattern recognition.
\end{IEEEbiography}

\begin{IEEEbiography}[{\includegraphics[width=1in,height=1.25in,clip,keepaspectratio]{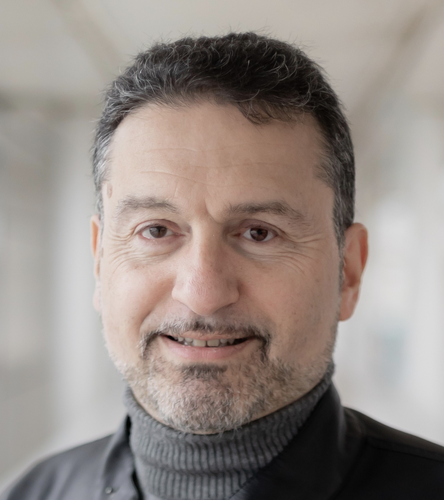}}]{Laurent Najman}
received the Habilitation {\`a} Diriger les Recherches in 2006 from University of Marne-la-Vall{\'e}e, a Ph.D. of applied mathematics from Paris-Dauphine University in 1994 with the highest honor (F{\'e}licitations du Jury) and an ``Ing{\'e}nieur'' degree from the Ecole des Mines de Paris in 1991. After earning his engineering degree, he worked in the central research laboratories of Thomson-CSF for three years, working on some problems of infrared image segmentation using mathematical morphology. He then joined a start-up company named Animation Science in 1995, as director of research and development. The technology of particle systems for computer graphics and scientific visualization, developed by the company under his technical leadership received several awards, including the ``European Information Technology Prize 1997'' awarded by the European Commission (Esprit programme) and by the European Council for Applied Science and Engineering and the ``Hottest Products of the Year 1996'' awarded by the Computer Graphics World journal. In 1998, he joined OCÉ Print Logic Technologies, as senior scientist. He worked there on various problems of image analysis dedicated to scanning and printing. In 2002, he joined the Computer Science Department of ESIEE Paris, where he is a Professor and a member of the Laboratoire Informatique Gaspard Monge (LIGM), Universit{\'e} Gustave Eiffel -- France. Since August 2025, he has been a Full Professor in the Department of Mathematics at Khalifa University -- UAE, where he is currently on leave from the LIGM. His current research interest is discrete mathematical morphology and discrete optimization.
\end{IEEEbiography}
\EOD

\end{document}